\documentclass[times,twocolumn,final,authoryear]{elsarticle}

\usepackage{arxiv}
\usepackage{framed,multirow}
\usepackage{amsmath}
\usepackage{amssymb}
\usepackage{latexsym}
\usepackage{multicol}
\usepackage{multirow}
\usepackage{makecell}
\usepackage{subcaption}
\usepackage{framed,enumitem}  
\usepackage{algorithm2e}
\usepackage{verbatim}
\usepackage{booktabs}
\usepackage{makecell}
\usepackage{array,ragged2e}
\usepackage{boldline}
\usepackage{setspace} 
\usepackage{pifont}
\usepackage{cancel}
\newcommand{\cmark}{\ding{51}}%
\newcommand{\xmark}{\ding{55}}%
\usepackage{url}
\usepackage{xcolor}
\usepackage{bibentry}
\usepackage{fancyhdr}
\definecolor{newcolor}{rgb}{.8,.349,.1}
\journal{Computer Vision and Image Understanding}

\begin{document}

\thispagestyle{empty}
\newcommand{\etal}{\textit{et al.}}
             
\begin{frontmatter}

\title{Co-segmentation Inspired Attention Module for Video-based Computer Vision Tasks}

\author[1]{Arulkumar \snm{Subramaniam}\corref{cor1}} 
% \cortext[cor1]{Corresponding author: 
%   Tel.: +91-9865084034;}
% \ead{aruls@cse.iitm.ac.in}
\author[1]{Jayesh \snm{Vaidya}}
\author[1]{Muhammed Abdul Majeed \snm{Ameen}}
\author[1,2]{Athira \snm{Nambiar}}
\author[1]{Anurag \snm{Mittal}}

\address[1]{Department of Computer Science and Engineering, Indian Institute of Technology Madras, IIT P.O., Chennai, Tamil Nadu - 600036, India}
\address[2]{Department of Computational Intelligence,
SRM Institute of Science and Technology, Chennai, Tamil Nadu - 603203, India}

\received{1 May 2013}
\finalform{10 May 2013}
\accepted{13 May 2013}
\availableonline{15 May 2013}
\communicated{A. Subramaniam}

\begin{abstract}

Video-based computer vision tasks can benefit from estimation of the salient regions and interactions between those regions. Traditionally, this has been done by identifying the object regions in the images by utilizing pre-trained models to perform object detection, object segmentation and/or object pose estimation. Although using pre-trained models is a viable approach, it has several limitations in the need for an exhaustive annotation of object categories, a possible domain gap between datasets and a bias that is typically present in pre-trained models. 
In this work, we propose to utilize the common rationale that a sequence of video frames capture a set of common objects and interactions between them, thus a notion of co-segmentation between the video frame features may equip the model with the ability to automatically focus on task-specific salient regions and improve the underlying task's performance in an end-to-end manner. In this regard, we propose a generic module called ``Co-Segmentation inspired Attention Module'' (COSAM) that can be plugged in to any CNN model to promote the notion of co-segmentation based attention among a sequence of video frame features. We show the application of COSAM in three video-based tasks namely: 1) Video-based person re-ID, 2) Video captioning, \& 3) Video action classification and demonstrate that COSAM is able to capture the task-specific salient regions in video frames, thus leading to notable performance improvements along with interpretable attention maps for a variety of video-based vision tasks, with possible application to other video-based vision tasks as well.

\end{abstract}


\end{frontmatter}

%\linenumbers

\vspace{-0.5cm}
\section{Introduction}
\label{sec:intro}
\vspace{-0.2cm}
Visual scene understanding from video data is an important research problem due to the large amount of videos generated and published everyday. Visual scenes captured in terms of a video typically focus on capturing one or more objects in terms of their activities and/or interactions between them. Thus, visual scene understanding is primarily aimed at localizing \& recognizing these objects and further determine interactions between them, if any. Several video-based computer vision tasks such as action recognition (\cite{simonyan2014two}), visual object segmentation \& tracking (\cite{ristani2016performance}), and video captioning (\cite{zhang2019object}) typically have such a scenario. To perform these tasks, earlier works used hand-crafted visual descriptors such as 2D/3D HoG (\cite{dalal2005histograms}), SIFT (\cite{lowe2004distinctive}) or  Optical flow (\cite{horn1981determining,bruhn2005lucas}) to aggregate visual information from the video frames, followed by a classification (or) regression layer according to the task at hand. More recently, the availability of large-scale datasets and computational resources have given rise to deep learning-based methods that have shown superior performance to hand-crafted features based approaches in several of these tasks. 

Deep learning approaches perform extremely well on the tasks that have extensive labeled samples available for training. Though annotated datasets are available for various tasks, not all datasets have fine-grained annotations (\textit{e.g.} object bounding boxes, segmentation masks or salient regions)  to enable the deep network to focus on important portions of image features. A straightforward approach to get fine-grained annotations and identify the object regions is to utilize pre-trained models to perform object detection, object segmentation and/or object pose estimation. However, such an approach has several limitations:  First, the object categories in the pre-trained models' training dataset may not cover all the object categories exhaustively needed for general computer vision tasks. Thus the object-related cues from the pre-trained models may prove to be incomplete for the underlying tasks. For instance, object detectors may not cover the objects belonging to the tail distribution such as star fish or baseball bat that may be present in tasks such as video captioning and video action classification. Further, while the pose estimation (or) pedestrian detection / segmentation models can effectively locate the person's key joint locations (\textit{e.g.} head, torso and legs), they miss out on the salient accessories associated with the subject (\textit{e.g.} backpack, bag, hat and coat) that are also important cues for tasks such as video captioning and video-based person re-identification (Re-ID). 

Second, the deep networks are shown to be susceptible to a domain difference between the data distributions of the training and target tasks. Learning domain-independent features is a research problem of its own. Owing to this, the domain gap between the pre-trained models' training dataset and target task's dataset may negatively impact the performance of the underlying task. For instance, in the case of Person Re-ID, pedestrian detector / segmentation methods and pose estimation methods are trained using frames from human activities and taken from YouTube or sports activity videos.  Hence, they may not cover the drastic viewpoint variations in surveillance scenarios, \textit{e.g.} a top-view. Also, surveillance images may not have a sufficient resolution for stable pose estimation. For the tasks of video captioning and video-action classification, the object detector / segmentation methods are pre-trained using MS-COCO and PASCAL VOC datasets. In a realistic setting, the data distributions of captioning (MSR-VTT, MSVD) and action classification (HMDB51, UCF101) are different from those pre-trained models' training data distribution and the performance may be influenced by this domain gap.

Thirdly, the bias present in pre-trained models may propagate to the target task leading to an inadvertently biased target model. The deep networks exhibit different types of biases due to factors such as the background, color, race (\cite{gwilliam2021rethinking}), gender (\cite{tang2021mitigating,zhao2017men}), context (\cite{singh2020don}), co-occurrence (\cite{petsiuk2021black}), spatial noise, dataset (\cite{tommasi2017deeper}) and object-size (\cite{nguyen2020evaluation}) etc. For instance, \cite{petsiuk2021black} show that the object detectors are vulnerable to learning the co-occurrence of an unrelated adversarial marker. Such biases may lead to a model that doesn't generalize well in the target task.
Added to these downsides, the pre-trained models are affected by nuisance variables such as background clutter, pose/viewpoint, illumination variations, and occlusions.

In this work, we present a method that does not have many of these
limitations.  We exploit the assumption that the sequence of frames belonging to a video often capture a (set of) common object(s) and interactions among them. Specifically, we propose to supplement the deep network with the ability to attend to the common set of task-specific salient regions between frames and hypothesize that this may help the network to improve the performance in an end-to-end manner. In this regard, we propose a novel generic attention module namely "Co-Segmentation Inspired Attention Module (COSAM)" for video-based applications. In particular, our proposed COSAM module facilitates intra-video attention inspired by the co-segmentation concept (\cite{vicente2011object, li2018deep}) by leveraging correlated information in multiple frames of the same video. COSAM is capable of performing soft attention on task-specific object regions depending on the performed vision task. In this work, we show the application of the proposed COSAM module in three major computer vision tasks: 1) Video-based person Re-ID, 2) Video captioning, and 3) Video action classification. Although we (\cite{subramaniam2019co}) originally proposed the idea towards video-based person Re-ID task, in this work, we extend the application domains to demonstrate the applicability of a generic "co-segmentation based attention" in various vision tasks.

We evaluate the effectiveness of the COSAM module by incorporating it into recent state-of-the-art models in these three major video-based computer vision tasks:

\begin{itemize}
    \setlength\itemsep{0.1em}
    \item \textit{Video-based person Re-ID: } Person re-identification (Re-ID) is the task of associating a person's image(s) among two or more camera views. It is posed as a fine-grained retrieval problem where the comparative scores between the query and the gallery instances are sorted to determine the best matched (top $k$) gallery instances for each query instance. In \cite{subramaniam2019co}, we proposed to incorporate the COSAM module into a typical video-based Re-ID pipeline (Sec. \ref{sec:videoreid_pipeline}) and demonstrated that the model performs superior to the state-of-the-arts. In this paper, along with \cite{subramaniam2019co}, we extend the experimental analysis to compare COSAM to the closely related non-local attention modules (\cite{wang2018non}) to compare and contrast them in terms of number of parameters, FLOPs and quantitative performance. 
    
    \item \textit{Video captioning:} Video captioning is the task of generating human understandable text about the given video automatically by capturing spatial \& temporal visual context along with a suitable mapping function to map the visual context into the text modality. In our work (\cite{vaidya2022co}), we consider the baseline as the state-of-the-art method (\cite{pan2020spatio}) that utilizes a pre-trained object detector (\cite{ren2015faster}) to obtain object-related features to aid video-captioning. We replace the object detector based \textit{local} branch by incorporating the COSAM module to automatically capture task-specific salient regions. Through extensive experiments, we evaluate the performance of the COSAM-based model and show that it is indeed possible to perform on par or better than state-of-the-art methods without using object detectors in the video captioning task.
    
    \item \textit{Video action classification:} Video classification is one of the fundamental research topics in Computer Vision and considered as a natural extension of the image classification task to videos,  \textit{i.e.} associate the video to one of the predefined classes. In this work, we make use of the recent state-of-the-art video classification model (\cite{kalfaoglu2020late}) as the baseline. Specifically, we demonstrate the significance of the COSAM module on this task by simply plugging it into the baseline 3D CNNs with two different backbones (I3D, ResNeXt-101) and show that COSAM is able to improve the performance of baseline architectures by capturing salient object regions in video frames.  
\end{itemize}

The paper is organized as follows. Section \ref{sec:relaredworks} details some of the related works, Section \ref{sec:cosegmodule} briefly describes the co-segmentation concept that our work is based upon and formulates a generic co-segmentation based attention module to activate common task-specific salient regions among a set of frames, Section \ref{sec:applications} illustrates the applicability of COSAM module on three video-based computer vision tasks, then Section \ref{sec:experiments} shows the experiments performed on these tasks along with a comparison with the state-of-the-art methods. Finally, section \ref{sec:conclusion} concludes with a summary and possible future directions.

\vspace{-0.4cm}
\section{Related works}
\label{sec:relaredworks}

\subsection{Evolution of attention modules}

In the early adoption of Deep Convolutional Neural Networks (DCNN) for image recognition, the model predominantly consisted of a sequence of convolution layers (typically convolution + BatchNorm + ReLU) followed by a linear classifier to classify the input (images / videos). For instance, AlexNet (\cite{krizhevsky2012imagenet}) uses 5 conv layers followed by 2 linear layers, VGG16 (\cite{simonyan2014very}) uses 16 conv layers followed by 2 linear layers, ResNet (\cite{he2016deep}) uses 5 conv blocks followed by average pooling and a linear classifier in which each conv block is made up of a sequence of conv layers with residual connections. 

\begin{table}[!ht]
\centering
\vspace{-0.2cm}
\begin{tabular}{|l|l|}
    \hline
    \multirow{6}{*}{\makecell[l]{Explicit feature\\ activation}}
    & SCA-CNN (\cite{chen2017sca}) \\
    &  SE-ResNet (\cite{hu2018squeeze}) \\
     &  BAM (\cite{park2018bam}) \\
    &  CBAM (\cite{woo2018cbam})  \\
    & DiverseAttention (\cite{li2018diversity}) \\
    & \textbf{COSAM (Ours)} \\\hline
    \multirow{5}{*}{\makecell[l]{Correlation and\\ feature \\aggregation}} &  ShowAttendTell (\cite{xu2015show}) \\
    & SelfAttention (\cite{vaswani2017attention}) \\
    &  NonLocal (\cite{wang2018non}) \\
    &  BERT (\cite{devlin2018bert}) \\
    & CrissCrossAttention (\cite{huang2019ccnet}) \\
    \hline
\end{tabular}
\vspace{-0.2cm}
\caption{Evolution of attention methodologies in the literature}
\label{tab:attention_evolution}
\vspace{-0.2cm}
\end{table}

Recent works try to improve the representation learning ability of backbone networks by augmenting it with ``attention'' blocks (spatial, channel, temporal) to improve representation learning. Based on the nature of attention, we broadly categorize these methods into two types namely, 1) \textit{Explicit feature activation}, and 2) \textit{Correlation and feature aggregation methods} as shown in the Table \ref{tab:attention_evolution}. 

\textit{Explicit feature activation (EFA)} based methods strive to estimate an ``importance'' mask (typically $\sim$[0, 1]) denoting the importance of the input features and use these masks to reweigh the input features. The motivation is to learn to boost the features that are important for the task and suppress the noisy features in an end-to-end manner via back-propagation. For instance,  SE-ResNet (\cite{hu2018squeeze}) proposed to learn a channel attention from global average pooled (GAP) feature maps. 
In a similar spirit, the Block Attention Module (BAM) (\cite{park2018bam}) proposed to learn spatial attention and channel attention masks at the bottleneck layers of ResNet. Different from BAM, Convolutional Block Attention Module (\cite{woo2018cbam}) proposed to use global statistics of feature maps such as average and max pooled channel descriptors to learn channel attention, followed by spatial attention. \cite{li2018diversity} proposed to learn diverse attention masks using multiple branches to focus on salient regions for image-based Re-ID. 
\textcolor{black}{SCA-CNN (\cite{chen2017sca}) proposed to estimate the spatial \& channel attentions for the image captioning task by conditioning on the LSTM’s previous hidden state and visual features.} Our work aligns with this paradigm of explicit feature activation, where the COSAM module learns to estimate spatial and channel importance (soft attention) masks conditioned on a correlation cost volume between the current frame and a set of reference frames.

On the other hand, \textit{Correlation and feature aggregation (CFA) methods} estimate the mutual attention weights among the set of input features (\textit{e.g.} word-level features in Natural language processing (NLP), pixel-level features in a 2D image and frame-level features in a video) and then, for each input feature, the output feature is estimated by a weighted sum of input features with attention weights. \cite{xu2015show} is one of the earlier works in this paradigm for the task of image captioning where at each time step of caption generation, the input context is gathered using a CFA construct between the hidden layer features of the last time step and the image features. Extending this style of feature interaction, \cite{vaswani2017attention} formulated a self-correlation and feature aggregation method called ``self-attention'' for NLP tasks. In \cite{vaswani2017attention}, three projections are obtained from each set of input features namely, \textit{query, key, value}. Then correlation is carried out between the query and key to get attention weights, followed by a feature aggregation step to aggregate the weighted sum of values for each input feature based on the attention weights. BERT (\cite{devlin2018bert}) follows the self-attention paradigm to learn a language model by predicting missing words in a sentence.
For video-based tasks, Non-local attention module networks (NLM) (\cite{wang2018non}) proposed to \textit{aggregate} the frame features based on cross attention probabilities between frame-wise pixel-level representations. The concept of COSAM is close to NLM, however COSAM proposes to explicitly \textit{activate} the frame-wise features at each pixel by learning spatial and channel attention weights instead of aggregating the features. By doing this, COSAM promotes feature interactions between multiple frames in an interpretable and computationally effective way in terms of the number of parameters and FLOPs (refer Sec. \ref{sec:ablation_reid} for more information).

\subsection{Object co-segmentation approaches}

Depending on the underlying algorithm, co-segmentation approaches can be grouped into two types: 1)  Graph-based (\cite{chang2015optimizing,jerripothula2016image,li2018unsupervised}), and 2) Clustering-based (\cite{joulin2012multi,tao2017image}). Graph-based methods exploit the shared structural representation of object instances from the given images to jointly segment the common objects, while the cluster based methods approach co-segmentation by grouping pixels/super-pixels to determine common object regions. Earlier methods (\cite{rother2006cosegmentation, vicente2011object}) used hand-crafted feature descriptors such as SIFT (\cite{lowe2004distinctive}) and HOG (\cite{dalal2005histograms}) to represent object instances, while  recent methods increasingly use deep learning approaches. 
Among the deep learning based methods, \cite{li2018deep} used a deep model to co-segment the regions based on the semantic similarity. \cite{hsu2018co} proposed to co-segment a particular category of objects in an unsupervised way by formulating contrastive objective functions to differentiate between foreground and background. Further, some methods (\cite{chen2018semantic, li2018deep}) proposed spatial and/or channel-wise attention-based bottleneck layers in an encoder-decoder architecture to activate semantically related features. Despite having a rich literature, the co-segmentation concept in other computer vision tasks is underexplored, and our work marks one of the first attempts exploring the application of co-segmentation to aid other vision tasks.

\subsection{Person re-identification}

Earlier approaches formulated the solutions by extracting hand-crafted features (\cite{bazzani2014sdalf,ma2012local}) followed by a suitable metric learning (\cite{bak2017one,prosser2010person,hermans2017defense}) for the re-identification task. Recent approaches utilize deep CNN models (\cite{ding2015deep,fu2018horizontal,xiao2017joint,hermans2017defense,zheng2018discriminatively}) to extract features and learn a suitable metric in an end-to-end manner. 
A common approach in many of the video Re-ID approaches is to extract frame-level features followed by temporal modelling / feature aggregation and classification, \textit{i.e.} LSTM/pooling (\cite{mclaughlin2016recurrent,zhou2017see,liu2018video}). For instance, \cite{mclaughlin2016recurrent} \& \cite{chung2017two} use deep CNNs to extract features from RGB \& optical flow and a recurrent layer followed by a temporal \textit{average} pooling (TP$_{avg}$) to aggregate features. These methods are easily affected by noisy background clutter, occluded objects and misalignment of the person that results in a drastic drop of Re-ID accuracy.

The solutions to overcome these nuisance factors broadly follow three types of approaches: 1) \textit{using pose estimation methods} to focus on the spatial region of the person and avoid features from the background (\cite{su2017pose}, \cite{suh2018part}), 2) \textit{using the segmentation mask of the subject \textit{i.e.}, person} from a pre-trained segmentation model (\cite{ren2015faster, he2017mask}) to avoid background clutter and extract features only from the foreground (\cite{qi2018maskreid,song2018mask}), and 3) \textit{using attention-driven approaches} to focus on salient regions in an end-to-end manner (\cite{li2018diversity,wang2016person}). 
Among these approaches, using pre-trained pose and segmentation models suffer from downsides such as the lack of ability to capture salient accessories and domain gap between original training \& re-ID datasets as mentioned in Section \ref{sec:intro}. 
Further, the attention models (\cite{zhou2017see,li2018diversity,wang2016person}) are  still sub-optimal since they work on ``per-frame" basis without using rich spatiotemporal cues in the video. Different from these methods, in our work, we utilize spatiotemporal features to co-segment the task-specific salient regions and activate them in an unsupervised way to aid the Re-ID task.

\subsection{Video captioning}

A typical methodology followed by deep learning methods for the video captioning task (\cite{pan2016jointly,venugopalan2015sequence,rohrbach2017movie,venugopalan2014translating}) is to extract global 2D/3D CNN abstract features (\cite{simonyan2014very,he2016deep,xie2017aggregated,carreira2017quo}) and then use a temporal modeling scheme (\textit{e.g.} an LSTM) to model the temporal relationship between the frames and then use a language decoder to generate the captions (\cite{hochreiter1997long,chung2014empirical,vaswani2017attention}).  Though the results from global 2D/3D CNN features are promising, they fail to capture local features needed for captioning. To gain fine-grained local cues, some methods (\cite{zhang2019object,pan2020spatio,zhang2020object,ma2018attend,zhou2019grounded}) have used pre-trained object detectors (\cite{ren2015faster,he2017mask,redmon2016you}) to extract local object features and shown a significant performance improvement. 
Following this paradigm, one of the recent methods (\cite{pan2020spatio}) proposed a two-branch network where a ``global scene branch'' extracts global features using 2D and 3D CNNs and a ``local branch'' extracts object specific features using a pre-trained object detector.

Though object detectors support video-captioning models with prior information on objects, these models are limited by the unavailability of exhaustively-annotated object categories in the detection datasets and domain gap between detection \& captioning datasets as highlighted in Section \ref{sec:intro}. Thus, the performance of the captioning task may be capped by the performance of pre-trained models in the video captioning dataset.  
In our work, we explore an alternative solution to eliminate the use of object detector by equipping the model with the COSAM module to learn and focus on task-specific salient regions in an end-to-end manner.

\subsection{Video  classification}

A typical pipeline for video classification task involves three core components: 1) {A feature extractor}, 2) Temporal modeling layer and 3) Classification layer. The \textit{feature extractor} is used to capture abstract features from the given video frames and can consist of a 2D (\cite{krizhevsky2012imagenet,he2016deep}) or 3D CNN with one (\cite{karpathy2014large}) or multiple branches (\cite{simonyan2014two}) depending on the nature of the input (RGB and/or optical flow). Further, the \textit{temporal modeling layer} learns temporal interactions between frame-wise features and summarizes the features. Naive pooling (\cite{wang2016temporal}), recurrent aggregation (\cite{wu2015modeling}) or RNN/LSTM layers (\cite{donahue2015long}) are some of the prevalent temporal modeling layers. Finally, the \textit{classification layer}  takes the summarized features as input and classifies into a set of predefined categories. Typically a set of linear layers with output nodes equal to the number of classes is used as a classification layer.

Initial attempts (\cite{yue2015beyond,donahue2015long}) to find better temporal modeling schemes either used a naive temporal average pooling or RNN/LSTM layers to capture the temporal information. Recent methods use attention mechanisms to model temporal dependency due to its strong empirical performance in other vision and NLP tasks. \cite{kalfaoglu2020late} used a BERT-style self-attention mechanism and pooling strategy to learn temporal modeling and showed promising performance improvement. In our work, we adopt \cite{kalfaoglu2020late} as our baseline network and attempt to study the effectiveness of the COSAM module in this method. 

Striving towards better \textit{backbone networks}, 3D CNNs have emerged as a natural choice considering its inherent nature to capture spatiotemporal information. Several works (\cite{yao2015describing,carreira2017quo,diba2017temporal}) utilize 3D CNNs to improve the performance by efficiently capturing spatiotemporal context from both appearance (RGB) and motion (optical flow) data. Recent methods employ attention schemes (self-attention / cross-attention) (\cite{xu2015show,girdhar2019video,chi2020non,wang2018non}) and Graph Convolutional Networks (GCN) to capture rich contextual information between video frames. For instance, \cite{barrett2015action} proposed to automatically identify the most discriminative temporal sub-sequences in a video. \cite{wang2018non} introduced non-local blocks to aggregate the features from video frames by means of correlating frame features. In our work, we attempt to incorporate the COSAM module into a recent state-of-the-art CNN (\cite{kalfaoglu2020late}) to automatically learn task-specific salient regions for video classification by means of co-segmentation inspired attention.

\section{Co-segmentation inspired attention module (COSAM)}
\label{sec:cosegmodule}

\subsection{Concept of object co-segmentation}
\label{sec:coseg_concept}

\begin{figure}[!ht]
\vspace{-0.2cm}
\begin{center}
\includegraphics[width=0.65\linewidth]{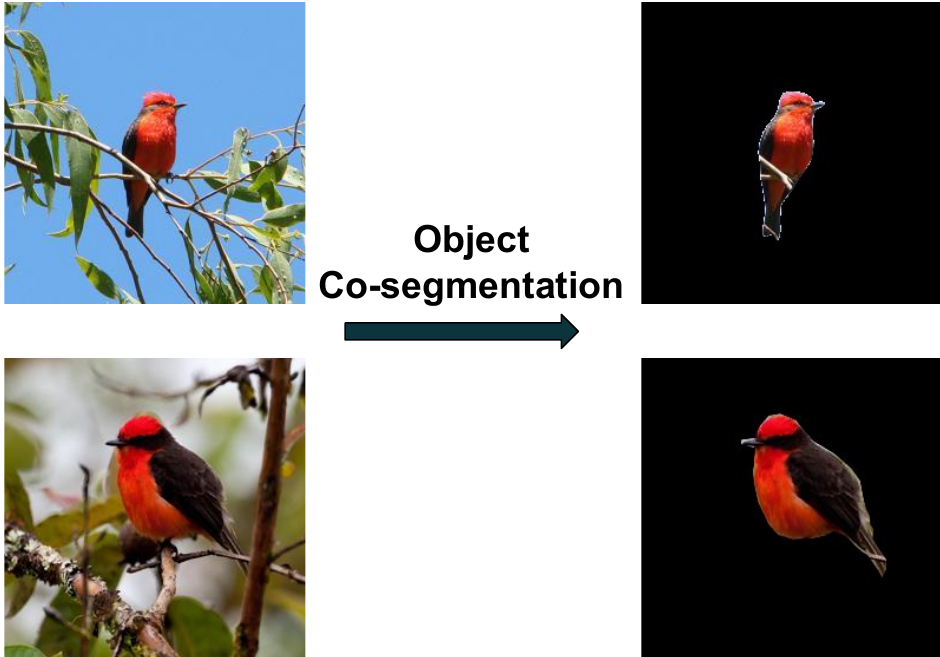}
 \vspace*{-0.3cm}
\caption{An example illustration of object co-segmentation using images from the Caltech-UCSD Birds 200 (\cite{WelinderEtal2010}) dataset.
\label{fig:coseg-sample}} 
\end{center}
\vspace*{-0.7cm}
\end{figure}

Object co-segmentation is the task of identifying and segmenting common objects from two or more images according to ``some'' common characteristics (\cite{vicente2011object, li2018deep}) such as the similarity of the object-class and appearance. An illustration of co-segmentation is depicted in Fig.~\ref{fig:coseg-sample}. 

Inspired by this task, we encompass an assumption that a sequence of frames belonging to a video often capture a common set of objects and interactions between them.
Thus, learning to attend to common task-specific salient regions between frames in an end-to-end manner may aid a deep network to improve the underlying task's performance.
To achieve this, we propose to incorporate some common saliency associated with the captured common object(s) among video frames that can enhance the features from the objects and suppress irrelevant background features. 
In this regard, we propose a novel Co-Segmentation inspired Attention Module (COSAM) that can be plugged-in  into a variety of CNN models.

\textcolor{black}{While the conventional co-segmentation task (\cite{vicente2011object, li2018deep}) aims to explicitly segment the common objects \textit{via hard segmentation masks}, the proposed COSAM is a \textit{soft attention} mechanism inspired by the co-segmentation concept to activate common task-specific salient regions among video frame features. In particular, COSAM induces the \textit{soft attention} on spatial and channel dimensions through a set of learnable parameters acting on correlative information (\textit{e.g.,} spatial correlation cost volume, common channel activations) between video frame features. These learnable parameters enable COSAM to adaptively attend to common salient regions depending on the underlying task \& its objective functions.}

\begin{figure*}[!ht]
\vspace{-0.4cm}
\centering
\includegraphics[width=\textwidth]{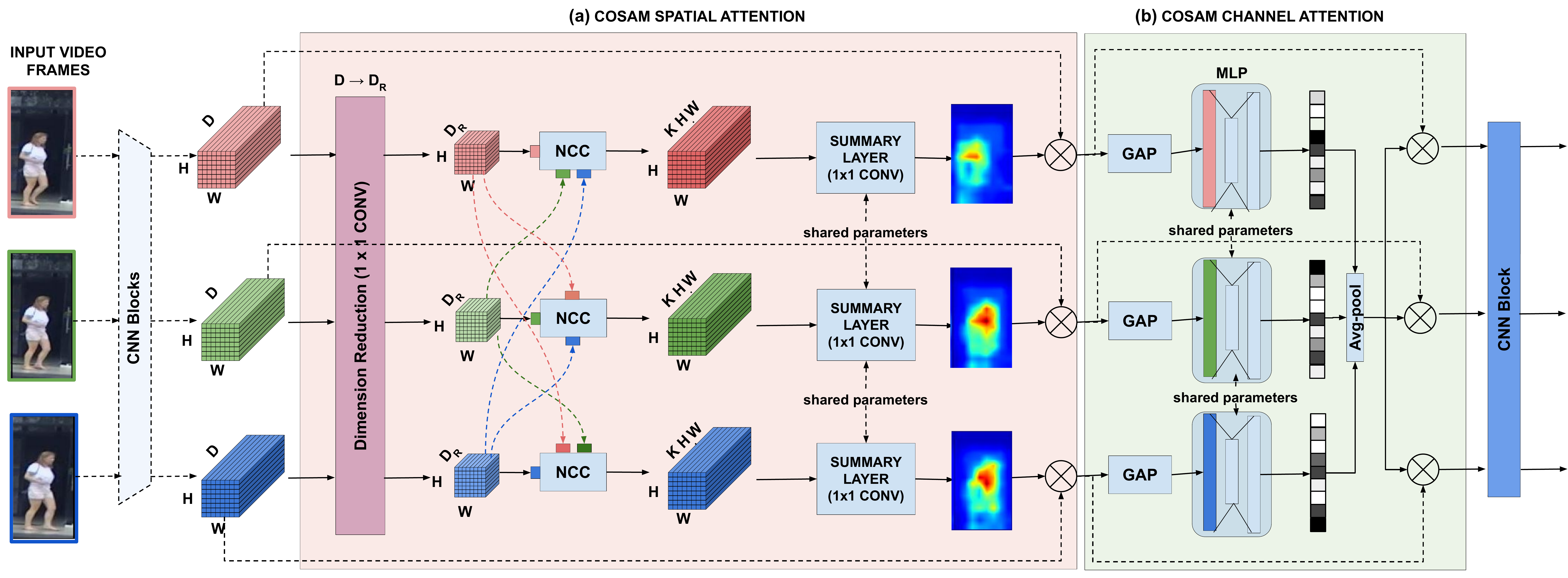}
\vspace{-0.7cm}
\caption{Illustration of Co-Segmentation inspired Attention Module (COSAM). The exemplary images are taken from the MARS (\cite{zheng2016mars}) person re-identification dataset.  (a) In the \textbf{COSAM spatial attention} step, the dimensionality reduced ($D \rightarrow D_R$) feature maps are passed
through a Normalized Cross Correlation (NCC) matching layer and a summarization layer to get the corresponding spatial attention mask.
(b) In the COSAM channel attention step, the common channel characteristics are used to activate the common features and suppress the noisy features.
\label{fig:cosam}}
\vspace{-0.4cm}
\end{figure*}

We briefly review two deep networks in the literature for co-segmentation that have inspired our work. \cite{li2018deep} proposed a siamese architecture with an encoder and a decoder to co-segment the common objects in a supervised learning framework. In this process, two or more images are passed through the encoder to get abstract feature maps. Then, in the bottleneck layer, mutual correlations of spatial feature descriptors are calculated to measure the commonality between the image features and this correlation based cost matrix is further passed to the decoder to estimate the co-segmentation mask. The same work also mentioned the idea of group co-segmentation to handle a group of images simultaneously. In another work, \cite{chen2018semantic} explored co-segmenting images by conditioning on common channel activations. To achieve this, they proposed a Siamese encoder-decoder architecture in which the channel activations of one image are conditioned on the channel activations of the other image (in image-pairs) in the bottleneck layer. For co-segmenting among a group of images, the channel activations are conditioned by taking the average channel activations of all the images. Our COSAM layer is inspired by the group co-segmentation approach from both of these papers, but is reformulated for video-based tasks.

The COSAM module takes the frame features $\{F_n\}_{n=1}^N$ of a video $V$ with dimension $N \times D \times H \times W$ as input and outputs the activated features of the same dimension as of the input ($N \times D \times H \times W$). Here, $N = $ number of input frames of the video, $D = $ number of channels in the feature maps, $H = $ height, and $W = $ width respectively. Further, the COSAM module is carefully designed to make it convenient to be plugged-in inside any deep CNN.
The functionality of the COSAM module illustrated in Fig. \ref{fig:cosam} consists of two steps, namely: 1) Spatial attention, and 2) Channel attention.

\subsection{COSAM Spatial Attention}
\label{sec:spatialatt}
In the \textit{spatial attention step} (Fig. \ref{fig:cosam}a), the goal is to estimate a spatial mask for each frame that only activates the salient spatial locations consulting with a set of reference frames. To achieve this, first the features of the input frames with dimension $N \times D \times H \times W$ are passed through a dimensionality reduction layer consisting of a $1 \times 1$ convolution layer followed by BatchNorm (\cite{ioffe2015batch}) and ReLU (\cite{nair2010rectified}) to perform dimensionality reduction ($D \rightarrow D_R$, $D_R << D$). Dimensionality reduction is specifically carried out to reduce the computational overhead. Next, to activate salient locations in the feature maps, we utilize the concept of co-segmentation (\cite{li2018deep}) among the dimensionality reduced features to induce the notion of common objects. Specifically, for each frame's feature map with dimension $D_R \times H \times W$, we consider the channel-wise feature vector at every spatial location $(i, j)$ ($1 \le i \le H, 1 \le j \le W$) as a $D_R$ dimensional local descriptor of the frame at location $(i, j)$, denoted by $F_{n}^{(i, j)}$. Further, for each feature map $F_n$, we select $K$ reference feature maps $\{R_k\}_{k=1}^K$ ($1 \le K \le (N-1)$) from the remaining frames. To match the local regions across frames, for each frame feature map $F_{n}$ and its location $(i, j)$, we compare the local descriptor  $F_{n}^{(i, j)}$ with all the local descriptors of its corresponding reference feature maps $R_k^{h, w}$ ($1 \le k \le K, 1 \le h \le H, 1 \le w \le W$) to get a cost volume of dimension $(KHW) \times H \times W$. Here, the comparison is carried out using Normalized Cross Correlation (NCC) between the local descriptors as it is robust to illumination variations and this was found to be more robust than a simple correlation (\cite{subramaniam2016deep}). The comparison results are reshaped into a 3D cost volume where each spatial location $(i, j)$ holds the comparison values. The idea of creating a cost volume in terms of matching the descriptors in an end-to-end learning framework has also been employed in other Computer Vision tasks such as geometric matching (\cite{rocco2017convolutional}), image-based Re-ID (\cite{subramaniam2016deep}) and stereo matching (\cite{kendall2017end}) among others.

\begin{figure*}[!th]
\vspace{-0.4cm}
\begin{center}
 \includegraphics[height=120px,width=0.9\linewidth]{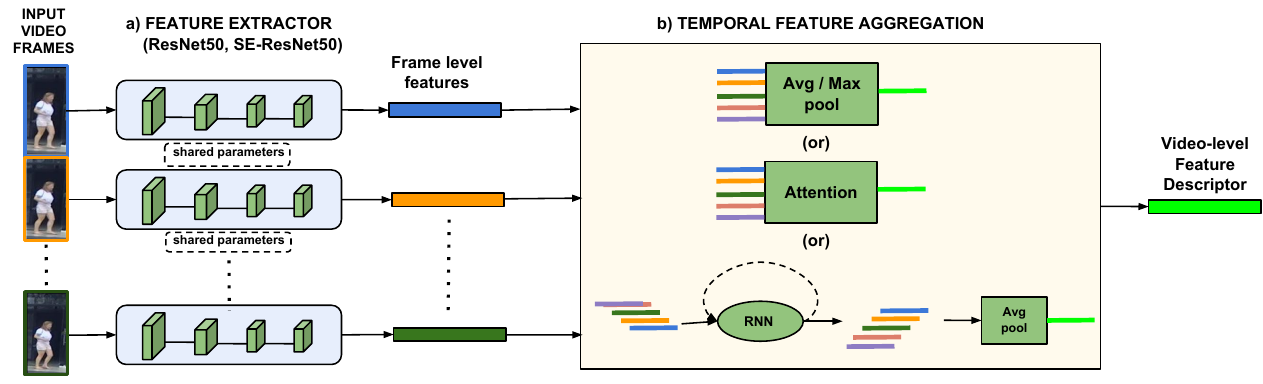}
\end{center}
\vspace{-0.6cm}
\caption{A standard Video-based Re-ID framework \label{fig:video-reid-pipeline} containing (a) a Feature extractor and (b) Temporal feature aggregation components.}
\vspace{-0.1cm}
\end{figure*}

\begin{figure*}[!ht]
\vspace{-0.3cm}
\centering
\includegraphics[width=\textwidth]{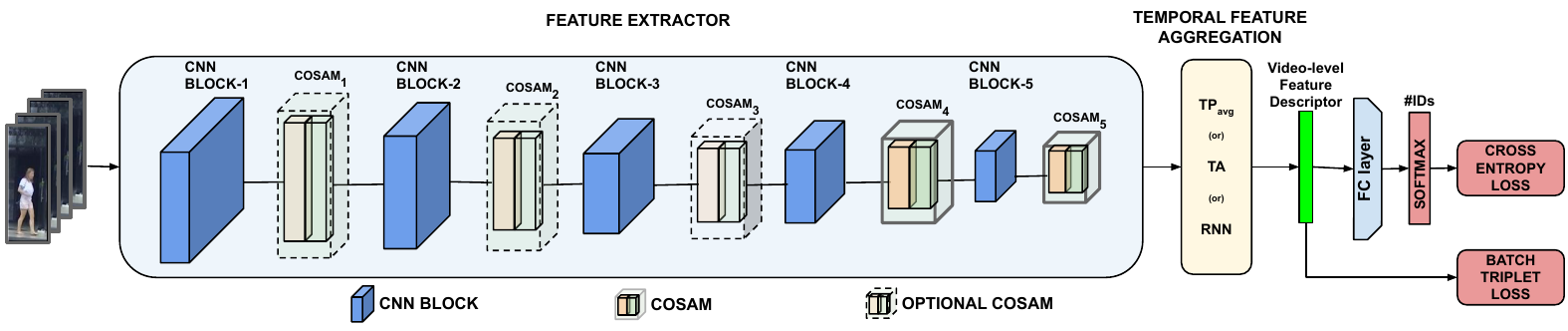}
\vspace{-0.6cm}
\caption{COSAM-based model architecture for the task of video-based person re-identification. COSAM modules are inserted between consecutive ResBlocks to induce a notion of co-segmentation based attention.
\label{fig:reid_cosam}}
\vspace{-0.6cm}
\end{figure*}

Mathematically, cost volume creation can be defined as:
% \vspace*{-0.3cm}
\begin{equation}
\begin{split}
\text{Cost volume}_{(n)}(i, j) =& \{ NCC\left(F_{n}^{(i, j)}, R_{k}^{(h, w)}\right) \\
& 1 \le k \le K, 1 \le h \le H, 1 \le w \le W \}.
\end{split}
\end{equation}

Given two descriptors $P, Q$ of $D_R$ dimension, the NCC operation is defined as:
% \vspace*{-0.3cm}
\begin{equation}
\begin{split}
\text{NCC}(P, Q) &=\frac{1}{D_R}\frac{\sum_{k=1}^{D_R} (P_k-\mu_P).(Q_k-\mu_Q)}{\sigma_P.\sigma_Q}.
\end{split}
\end{equation}

Here, $(\mu_P, \mu_Q)$ denote the mean of the descriptors $(P, Q)$, \& $(\sigma_P, \sigma_Q)$ denote the standard deviations of the descriptors $(P, Q)$ respectively. (A small value of $\epsilon=1e^{-4}$ is added to $\sigma's$ to avoid numerical instability). 

The cost volume of each frame is summarized using a $1 \times 1$ convolution layer followed by a sigmoid activation resulting in a spatial mask (dimension $1 \times H \times W$) for the corresponding frame. The spatial mask is multiplied with the corresponding frame's original input features $F_{n}$ to activate only the local regions of images that are in consensus with its reference frames. The output features after the spatial attention step are passed on to the channel attention step.

\subsection{COSAM Channel Attention}
\label{sec:channelatt}

In the channel attention step (Fig.~\ref{fig:cosam} (b)), the goal is to boost the common important channels between the frames and suppress the noisy channels. To achieve this, inspired by (\cite{chen2018semantic,hu2018squeeze}), we apply a Global Average Pooling (GAP) layer on the spatially-attended feature maps from the spatial-attention step and get a 1D feature of dimension $D$ for each frame. This resulting feature vector is passed through an Encoder-decoder style Multi-Layer Perceptron (MLP) followed by sigmoid activation to get a $D-$dimensional vector signifying the importance of each channel for every frame. The obtained channel importance vectors of all the $N$ frames are average pooled together on each channel dimension to estimate the global channel-importance. In essence, this step boosts the channels that are important across all the frame features and suppresses the noisy channels. Next, the averaged channel importance vector is multiplied with spatially-attended features to obtain the importance-weighted channel activations that are passed to the next layer.

\section{Applications of COSAM}
\label{sec:applications}

COSAM is carefully designed as a generic module to be plug-and-play compatible within the CNN models in various video-based computer vision tasks. In this work, we show the applicability of COSAM on three exemplary computer vision tasks namely: 1) \textit{Video-based person re-identification}, 2) \textit{Video captioning}, and 3) \textit{Video action classification}.
Specifically, we study the COSAM module by incorporating it into state-of-the-arts methods for these tasks.
The following sections (\ref{sec:personreid}, \ref{sec:captioning}, and \ref{sec:videoclassify}) describe the baseline architectures for each of these tasks and the modifications carried out to incorporate COSAM into the baseline architectures.

\subsection{Video-based person re-identification}
\label{sec:personreid}

In this section, we outline the integration of the COSAM module into a video-based person re-identification pipeline.

\subsubsection{Typical video-based Re-ID pipeline \label{sec:videoreid_pipeline}}

We follow a recent line of research yielding a strong baseline in video-based Re-ID that can be summarized into a template framework as shown in Fig.~\ref{fig:video-reid-pipeline}. It consists of two primary components : \textbf{ (a)  A Feature extraction network:} This is capable of extracting a meaningful abstract spatial representation from video frames either by hand-crafted features (SIFT, LBP, HoG, etc.) or automatically extracted deep CNN features by using pre-trained ImageNet models (\cite{he2016deep, hu2018squeeze}) such as ResNet and SE-ResNet.
\textbf{ (b) Temporal feature aggregation:} Here, the extracted frame-level features are aggregated to form a video-level feature vector to represent the person identity in the video. The complexity of feature aggregation techniques in the literature varies from a \emph{simple} temporal pooling (TP$_{max/avg}$) operation (average/max pooling) to \emph{complex} temporal attention (TA) and recurrent layer (RNN) based aggregation (\cite{gao2018revisit}). The aggregated video-level feature vectors are then used to compare (using $L_2$ distance or a learned metric) against other video instances for matching and retrieval purpose. In Table \ref{tab:summary}, we give a summary of prior work in video-based Re-ID using the aforementioned framework.

\begin{table}[!ht]
\vspace{-0.2cm}
% \scriptsize
\footnotesize
\centering
\begin{tabular}{|l|l|l|}
\hline
\textbf{Literature work} \newline  & \textbf{\makecell[l]{Feature\\extractor}} & \textbf{\makecell[l]{Feature\\ aggregation}}\\
\hline
%& three-layer CNN&TP\\
\makecell[l]{RCN for Re-ID\\ (\cite{mclaughlin2016recurrent})} &Custom 3-layer CNN&RNN + TP$_{avg}$\\
\hline
\makecell[l]{Two Stream Siamese\\(\cite{chung2017two})} &Custom 3-layer CNN&RNN + TP$_{avg}$\\
\hline
\makecell[l]{Jointly attentive \\ spatiotemporal pooling\\(\cite{xu2017jointly})}& \makecell[l]{deep CNN + \\spatial pyramid\\ pooling}& Attentive TP\\
\hline
\makecell[l]{Comp. Snippet Sim.\\(\cite{chen2018video})}&ResNet-50& \makecell[l]{LSTM, Co-attentive\\ embedding}\\
\hline
\makecell[l]{Part-Aligned\\(\cite{suh2018part})}
&GoogLeNet&Bilinear pooling\\
\hline
\end{tabular}
\vspace{-0.2cm}
\caption{A collection of approaches in the literature that are following the video-based Re-ID pipeline shown in Fig.~\ref{fig:video-reid-pipeline}. \label{tab:summary}}
\vspace{-0.4cm}
\end{table}

%\end{comment}

In the following section, we describe the network architecture incorporated with the COSAM module following the template framework in Fig. \ref{fig:video-reid-pipeline}.

\subsubsection{Network architecture}

Modern state-of-the-art network architectures (ResNet50, SE-ResNet, etc.,) that are used as feature extractors in video-based Re-ID contain multiple consecutive \textit{CNN blocks}, in which the convolution layers are grouped according to the resolution of output feature maps: ResNet50 and SE-ResNet50 have five blocks (one initial convolution block followed by four consecutive Residual (or) Squeeze and Excitation (SE) residual blocks). We propose to plug-in the COSAM layer after the \textit{CNN blocks} in these network architectures. An illustration of the proposed network architecture along with the COSAM layer is shown in Fig.~\ref{fig:reid_cosam}. After getting the output of every \textit{CNN block}, the feature extractor employs a COSAM layer to co-segment the features and then the co-segmented features are passed to the next \textit{CNN block}. At the end of the feature extractor, the temporal aggregation layer (TP$_{avg}$ or TA or RNN) is applied to summarize the frame-level descriptors into a video-level descriptor. This video-level descriptor is used to predict the probability that the video belongs to a particular person identity. 

\vspace{-0.1cm}
\subsubsection{Objective functions}

For a fair comparison with the baseline (\cite{gao2018revisit}) and due to their suitability for our task, we use the same loss functions as in \cite{gao2018revisit}.
The overall loss function can be written as:
\vspace*{-0.2cm}
\begin{equation}
L = \sum_{i=1}^B \left\{ L_{CE} + \lambda L_{triplet}(I_i, I_{i_+}, I_{i_-}) \right\}.
\end{equation}
\vspace{-0.4cm}

\begin{figure*}[!ht]
\vspace{-0.2cm}
\begin{center}
\includegraphics[height=0.36\linewidth,width=\linewidth]{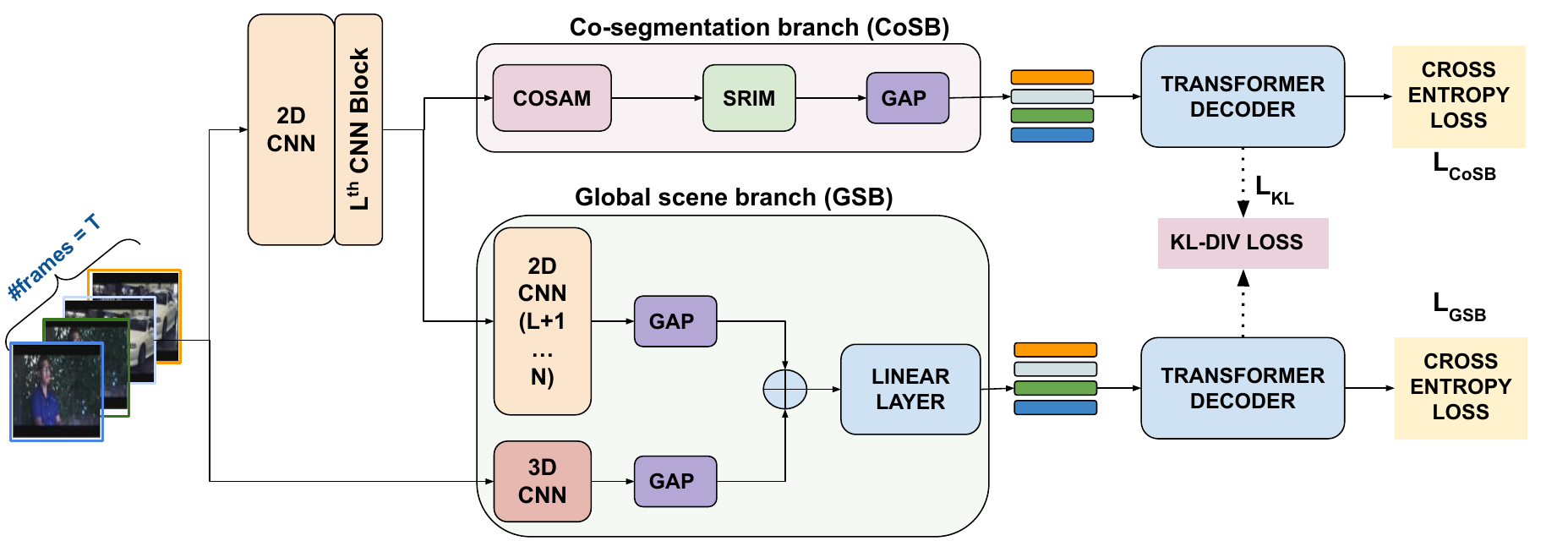}
\end{center}
\vspace{-0.6cm}
\caption{Illustration of the proposed model architecture for Video captioning task. The model consists of two branch viz., 1) Global Scene branch (GSB) - Sec. \ref{sec:globalscene}, and 2) Co-segmentation branch (CoSB) - Sec. \ref{sec:coatt}. Both branches have a separate Transformer as decoder (Sec. \ref{sec:decoder}). GSB captures the global scene by using 2D and 3D CNN global descriptors. CoSB attends to the salient regions via the COSAM module (Sec. \ref{sec:cosegmodule}) and promotes interaction among the salient regions through SRIM (Sec. \ref{sec:srim}). Generated captions are evaluated using standard cross entropy loss. Further, the salient region information is distilled from CoSB to GSB via KL-divergence loss. Here, GAP = Global average pooling, $\bigoplus$ = feature concatenation.\label{fig:arch}}  
\vspace{-0.1cm}
\end{figure*}

Here, $L_{CE}$ \&  $L_{triplet}$ refer to the cross-entropy loss and batch triplet loss respectively and $\lambda$ refers to the trade-off parameter between the losses (we use $\lambda = 1$, as per \cite{gao2018revisit}), $B$ = batch size \& $(I_i, I_{i_+}, I_{i_-})$ refer to the $i$th image in the batch and its hard positive and hard negative pair within the current batch, respectively.

\paragraph{Cross-Entropy loss ($L_{CE}$):} This supervised loss is used to calculate the classification error among the identities. The number of nodes in the softmax layer depends on the number of identities in the training set. 

\paragraph{Batch triplet loss ($L_{triplet}$): } To reduce the intra-class variation and to increase the inter-class variation, the training instances are formed as a triplet where each triplet contains an anchor, a positive instance that belongs to the same class as the anchor and a negative instance that belongs to a different class than the anchor. Hard negative mining is carried out on the fly in each batch to select the hardest examples that pose a challenge for the model. Let $\{f_{I_{A}}, f_{I_{+}}, f_{I_{-}}\}$ be the video-level descriptors of a triplet, where $I_{A}, I_{+}, I_{-}$ are the anchor, positive and negative examples respectively. The triplet loss function is defined as:
\vspace{-0.2cm}
\begin{equation}
% \begin{split}
L_{triplet}(I_{A}, I_{+}, I_{-}) = \max\left\{ D\left(f_{I_{A}}, f_{I_{+}}\right) 
% \\&\hspace{-0.5cm}
- D\left(f_{I_{A}}, f_{I_{-}}\right) + m, 0 \right\}.
% \end{split}
\end{equation}
Here, $m$ is the margin between the distances, $D(i, j)$ denotes the distance function between two descriptors $i, j$.

The Cross-entropy loss function is applied on the softmax probabilities obtained for the identities and the batch triplet loss is applied on the video-level descriptors to back-propagate the gradients. The experimental results along with a comparison with the state-of-the-art methods are shown in Section \ref{sec:experiments_reid}.

\subsection{Video captioning}
\label{sec:captioning}

In this section, we describe the application of the COSAM module to the task of video captioning. Object detectors are commonly used in the video captioning literature to obtain object-specific features. In our work, we propose to eliminate the usage of pre-trained object detectors from a state-of-the-art video captioning pipeline (\cite{pan2020spatio}) by augmenting it with COSAM. We hypothesize that the dependency on pre-trained object detectors could be avoided if the model is supplemented with the ability to focus on salient object regions automatically.

%-------------------------------------------------------------------------
\subsubsection{Co-segmentation aided two-stream architecture}

The overall model architecture is illustrated in Fig. \ref{fig:arch}. It follows a two-branch architecture: 1) \textit{Global scene branch (GSB)} for capturing global scene cues, 2) \textit{Co-segmentation branch (CoSB)} for capturing salient region features based on co-segmentation between frame-level spatial features. Each branch follows an encoder-decoder architecture. First, the input frames $\{F_i\}_{i=1}^T$ of a video $V$ with dimension $T \times 3 \times H \times W$ are passed through GSB and CoSB simultaneously to predict the captions individually. Here $F_i$ = $i^{th}$ frame, $T$ = number of frames, $3$ channels belong to RGB, $H$ = height, and $W$ = width of the frames respectively. Next, we employ the cross-entropy loss to evaluate the predicted captions from both the branches. Further, we employ a KL-divergence loss to impose a knowledge distillation constraint that requires both the branches to predict the same captions with similar confidence level. This ensures that the knowledge of learned salient regions from CoSB is propagated to GSB. \cite{pan2020spatio} follows a similar approach. However, they make use of an object-detection branch that requires a pre-trained object detector. In the following sections, we describe the constituting components of the model.

\begin{figure*}[!ht]
\vspace{-0.5cm}
\centerline{\includegraphics[width=\linewidth, height=0.23\linewidth]{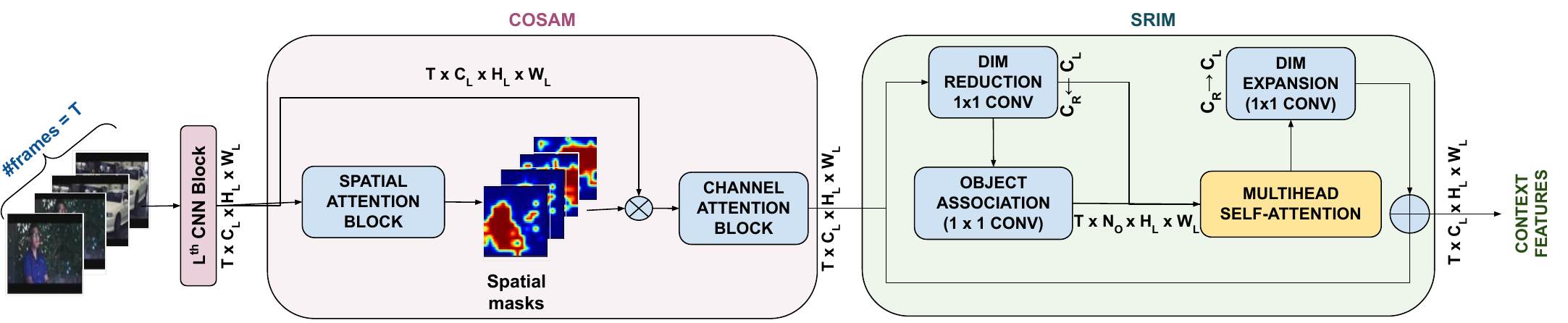}}
\vspace{-0.3cm}
\caption{Video captioning architecture's Co-Segmentation branch (CoSB) consists of two sub-modules COSAM and SRIM. 1) COSAM performs spatial and channel attention on the input feature map(s), 2) Next, Salient Region Interaction Module (SRIM) promotes information propagation between salient regions via multi-head self-attention mechanism. Here, $\bigotimes$ = point-wise multiplication, $\bigoplus$ = feature addition.  \label{fig:cosb}}
\vspace{-0.4cm}
\end{figure*}

\subsubsection{Global scene branch (GSB)}
\label{sec:globalscene}
The goal of GSB is to capture the global scene-level information to aid the captioning task. To achieve this, we utilize an ImageNet (\cite{deng2009imagenet}) pre-trained 2D CNN to capture frame-wise features and a Kinetics (\cite{kay2017kinetics}) pre-trained 3D CNN to capture temporal features of the video. Specifically, given the input frames $\{F_i\}_{i=1}^{T}$ of dimension $T \times 3 \times H \times W$, we pass the frames through a 2D ResNet-101 (\cite{he2016deep}) pre-trained on ImageNet to obtain frame-wise features of dimension $T \times 2048$ after global-average pooling (GAP) layer. The input frames are also passed through a 3D ResNeXt-101 (\cite{xie2017aggregated}) pre-trained on kinetics dataset to obtain temporal features of dimension $T \times 2048$ from its fully-connected layer before the classification layer. These per-frame 2D and 3D features are concatenated together and are projected to 512 dimensions to result in features $F_{GSB}$ of dimension $T \times 512$ for each video.

\subsubsection{Co-segmentation branch (CoSB)}
\label{sec:coatt}
Global-scene information captured by GSB (Sec. \ref{sec:globalscene}) may not be sufficient for the task of video captioning which requires fine-grained understanding of the video. Hence, different from GSB, the goal of CoSB is to focus on salient spatial regions of the frames that serves the task better. To this end, we propose to employ a COSAM module (Section \ref{sec:cosegmodule}) to focus on salient regions. Specifically, the CoSB branch (Fig. \ref{fig:cosb}) takes spatial feature maps of dimension $T \times C_L \times H_L \times W_L$ and uses a COSAM module to capture task-specific salient regions of frames, followed by a salient-region interaction module (SRIM) to capture interaction between those regions. Here, $C_L, H_L, W_L$ = number of channels, height, width of feature maps after $L^{th}$ CNN block. In our work, we obtain these input feature maps from 2D ResNet-101 used in GSB to share weights and reduce the model complexity.
We will briefly mention the architecture of the COSAM and SRIM modules in the following sections.

\paragraph{Integration of COSAM\label{sec:captioning-cosam-arch}} In the \textit{CoSB} branch, first we apply the COSAM module within adjacent frames' spatial feature maps to capture the salient regions. 
As demonstrated in Section \ref{sec:cosegmodule}, COSAM consists of two consecutive attention blocks namely: 1) \textit{Spatial attention block 
(COSAM-SAB) } to activate common spatial salient regions and suppress non-informative regions, 2) \textit{Channel attention block (COSAM-CAB) } to activate common informative channels. The design of COSAM-SAB and COSAM-CAB follows the architecture shown in Fig. \ref{fig:cosam}.
Specifically, the input feature maps of dimension $T \times C_L \times H_L \times W_L$ are passed through COSAM-SAB to generate a spatial attention mask of dimension $T \times 1 \times H_L \times W_L$. For each frame $F_i$, COSAM-SAB correlates every location's feature with its adjacent frame features (i.e., $F_{i-1}, F_{i+1}$)  and creates a cost volume. A summary convolution layer takes the cost volume as input and produces a spatial attention mask. The spatial mask is multiplied with the corresponding original features to obtain spatially refined features. Further, COSAM-CAB takes the spatially refined features as input, then through a GAP followed by an MLP, it produces channel attention weights ($T \times C_L$). These channel attention weights are multiplied with spatially refined features to output co-attended features of dimension $T \times C_L \times H_L \times W_L$.

\begin{figure*}[!ht]
\vspace{-0.3cm}
\centering
\includegraphics[width=\textwidth]{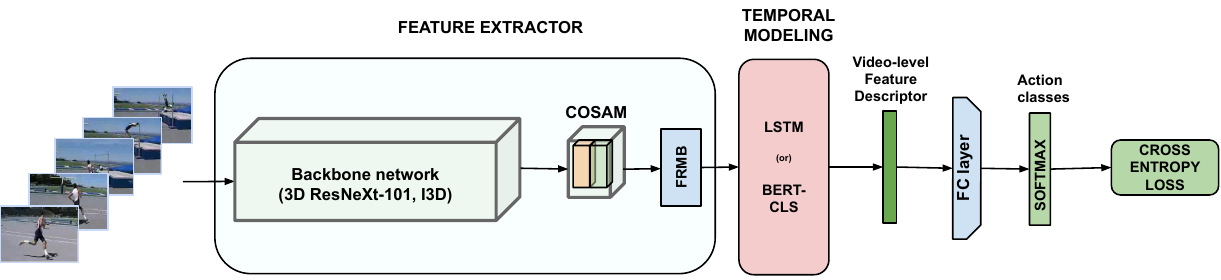}
\vspace{-0.6cm}
\caption{Illustration of the COSAM-based network architecture adopted for video action classification task. The baseline model (\cite{kalfaoglu2020late}) consists of single stream feature extractor (3D ResNext-101, I3D) accepting RGB frames (dimension $T \times 3 \times H \times W$) as input and passes it through a set of 3D CNN blocks, followed by Feature Reduction with Modified Block (FRMB) module (\cite{kalfaoglu2020late}), temporal modeling layer (LSTM  or BeRT-CLS) and a classification layer. COSAM module is inserted after the feature extractor and before FRMB module to study the improvements. A cross entropy loss is employed as the objective function during training.}
\label{fig:model_videoclassify}
\vspace{-0.4cm}
\end{figure*}

\paragraph{Salient-region interaction module (SRIM)\label{sec:srim}}
The COSAM module may have selected the salient regions containing potentially multiple objects. To promote interactions between individual object-like regions, we propose to use a self-attention based salient region interaction module inspired from GloRe (\cite{chen2019graph}).  It takes the co-attended spatial feature maps of dimension $T \times C_L \times H_L \times W_L$ from the previous step (COSAM) as input, and passes it through a $1 \times 1$ 2D convolution layer to perform dimension reduction on channels ($C_L \rightarrow C_R$, $C_R << C_L$) to reduce computations. Next, we assume that every frame has $N_o$ objects (in this work, $N_o$ is set to $5$) and each pixel may belong to one of these objects. To figure out the object association of pixels, the dimension-reduced feature maps (dimension $T \times C_R \times H_L \times W_L$) are passed through an object-association module to output  \textit{object-association maps $\{O_i\}_{i=1}^T$} that associates every pixel of the feature map to one of $N_{o}$ objects. We utilize a $1 \times 1$ 2D convolution with $N_{o}$ output channels to achieve this. Further, based on this object association map, we apply weighted average pooling (WAP) on the dimension-reduced feature maps to output object feature descriptors of size $T \times N_o \times C_R$. i.e., for each object, the model outputs a descriptor of size $C_R$. In the next step, we promote interactions between these object features using multi-head self-attention (\cite{vaswani2017attention}). Each frame $F_i$'s object features ($N_o \times C_R$) are restricted to interact with the previous, current and next frame's ($F_{i-1}, F_i, F_{i+1}$) object features to avoid noisy feature interactions (refer to Sec. \ref{sec:ablation_caption} for ablation studies).

Next, these self-attended object features are distributed back to pixel-space through a dimension expansion block ($C_R \rightarrow C_L$) consisting of a $1 \times 1$ 2D convolution followed by a reverse mapping module based on object association maps $\{{O_i}\}_{i=1}^T$. These redistributed features (dimension $T \times C_L \times H_L \times W_L$) are added to the original input features (dimension $T \times C_L \times H_L \times W_L$) to yield context-aware features $\{F_{{c}_i}\}_{i=1}^T$.
Next, we apply GAP on $\{F_{{c}_i}\}_{i=1}^T$ to get per-frame feature descriptors $F_{CoSB}$ (dimension $T \times C_L$) to pass them to the decoder for caption generation. 

\subsubsection{Caption generation decoder\label{sec:decoder}}

For every video instance, after obtaining its features from GSB ($F_{GSB}$) and CoSB ($F_{CoSB}$), we use two separate decoders to predict captions from $F_{GSB}$ and $F_{CoSB}$ respectively, similar to \cite{pan2020spatio}. The decoder transformer generates a word at every time-step by attending to the input features as well as previously generated words in the caption. The predicted caption from the decoders is evaluated using cross-entropy loss. Further, to transfer salient region-specific learning from CoSB to GSB, we impose a knowledge distillation based constraint that both CoSB and GSB shall generate the caption words with similar confidence (refer Sec \ref{sec:loss} for objective functions). During test time, we use the captions from the Global scene branch for evaluation, as followed in \cite{pan2020spatio}.

\subsubsection{Objective functions\label{sec:loss}}
We use standard cross entropy loss for evaluating captions generated from the transformer decoders. Further, to promote knowledge propagation of salient regions from CoSB to GSB, we apply an online knowledge distillation constraint between GSB and CoSB branches via KL-divergence loss similar to \cite{pan2020spatio}. Given the word vocabulary $W$ and probabilities of caption words generated from global-scene branch $P$,  co-segmentation branch $Q$, the knowledge distillation constraint is formulated using KL-divergence as follows:
\vspace{-0.3cm}
\begin{align}
D_{KL}(P\ ||\ Q) \ = \sum\limits_{x \in W} P(x) \log\left(\frac{P(x)}{Q(x)}\right).    
\end{align}

The overall loss function is given by,
\begin{equation}
L = L_{GSB}+ \lambda L_{CoSB}+ \lambda_{KL}L_{KL}\ .
\end{equation}
Here, $L_{GSB}$ and $L_{CoSB}$ are the individual cross entropy losses for GSB and CoSB branches respectively. $L_{KL}$ denotes KL-divergence between probability distributions of GSB and CoSB branch. $\lambda$ and $\lambda_{KL}$ are hyper-parameters. We perform experiments on two datasets (MSR-VTT (\cite{xu2016msr}) and MSVD (\cite{chen2011collecting})) and illustrate the results along with state-of-the-art comparison in Section \ref{sec:experiments_caption}.

\subsection{Video action classification}
\label{sec:videoclassify}

In this section, we utilize the COSAM module to perform the task of video classification. In our work, without loss of generality, we take a straightforward approach to adopt a recent state-of-the-art baseline method (\cite{kalfaoglu2020late}) with BERT based late temporal modeling for action recognition and simply insert the COSAM module into the baseline method to analyze the contribution of the COSAM module.

\subsubsection{Network architecture}

We briefly mention the network architecture of the baseline method (\cite{kalfaoglu2020late})  in the following paragraphs. The network architecture (Fig. \ref{fig:model_videoclassify}) consists of three core components as follows: 

\paragraph{1) \textit{Feature extractor}} It extracts abstract features $\{F_i\}_{i=1}^T$ from the given video frames $\{f_i\}_{i=1}^T$ of dimension $T \times 3 \times H \times W$ where $T = $ number of frames in the video, 3 channels signify RGB channels, $H = $ height, $W = $ width respectively. Specifically, in our work, we study the effect of the COSAM module with two different backbone networks namely: 1) 3D ResNeXt-101, and 2) I3D. \textit{3D ResNeXt-101} follows the architecture of ResNet-101 in which the 2D residual blocks are replaced by 3D inception-based residual blocks. \textit{I3D} follows the architecture of the Inception network in which the 2D inception modules are replaced by 3D inception modules. These networks are equipped with Feature Reduction with Modified Block (FRMB) (\cite{kalfaoglu2020late} at the end and pre-trained with Kinetics dataset. We plug-in the COSAM module after the final convolution block of the backbone network (\textit{i.e.}, before FRMB).

\paragraph{2)\textit{ Temporal modeling layer}} It is used to model temporal dependencies between frame-level features and to finally obtain a summarised video-level feature. We follow the baseline method (\cite{kalfaoglu2020late}) to experiment with two kinds of temporal modeling layers namely: 1) \textit{BERT based} and 2) \textit{LSTM} based.
In the \textit{BERT based} temporal modeling layer, the frame-wise features are added with the position-encoding and further concatenated with a \textit{CLS} token. These position-encoded features are passed through a sequence of self-attention layers to perform multi-head self-attention on the frame-level features. Finally, the output features corresponding to the \textit{CLS} token are considered as the summarized video-level representation and passed to the next layer. 
A long short-term memory (LSTM) layer takes as input the global-average pooled (GAP) frame-level features sequentially and forwards it through time. The output at the last time step is considered as the summarized \textit{video-level feature}. 
%Following \cite{kalfaoglu2020late}, we perform experiments of BERT layer with  

\paragraph{3)\textit{ Classification layer}} The summarised video-level feature from the temporal modeling layer (LSTM or BERT-CLS) is passed through a fully connected layer to output classification logits corresponding to action classes. 

\paragraph{Overall pipeline} The feature extractor takes an input of dimension $T \times 3 \times H \times W$ and passes it through a set of $L$ 3D convolutional (or 3D residual) blocks and results in output features $O_L$ of dimension $\frac{T}{S_T^L} \times D_L \times \frac{H}{S_H^L} \times \frac{W}{S_W^L}$. Here, $S_T^L, S_H^L, S_W^L$ are strides in time, height, \& width dimensions respectively. In the case of the baseline model, the output features $O_L$ from the backbone network are passed through the FRMB module to obtain the final convolutional features of the given video snippet. Then, global average pooling (GAP) is applied to the convolutional features to obtain frame-wise features. These frame-wise features are passed through a temporal modeling layer (BERT or LSTM) to model temporal dependencies and to get a video-level summarized feature. The video-level features are passed through the classification layer to result in class-specific scores for the classes in the dataset. In the COSAM-based model, the output features $O_L$ are passed through the COSAM module to activate the set of common features and then passed through the FRMB module for further processing.

\subsubsection{Objective functions}

We employ a simple softmax cross entropy loss among the video action classes as the objective function during training. Specifically, given the training data $\{\{f^i_k\}_{i=1}^T, t_k\}_{k=1}^N$ ($f_k$ = $k^{th}$ training video instance, $t_k$ = the target class one-hot vector, $T = $ number of frames, $N=$ number of videos), the class softmax probabilities $\{p^j_k\}_{j=1}^C$ are obtained from the classification layer.  Then, the cross entropy loss is given by:
\vspace{-0.2cm}
\begin{equation}
L = -\sum_{k=1}^N\sum_{c=1}^C I(c = t_k) \log(p_{k}^c).
\end{equation}
\vspace{-0.2cm}

Here, $I(.)$ denotes an indicator function, $C$ = number of classes. We perform experiments on two datasets (HMDB51 (\cite{kuehne2011hmdb}) and UCF101 (\cite{soomro2012ucf101})) and illustrate the results along with state-of-the-art comparison in Section \ref{sec:experiments_videoclassify}.

\section{Experiments and results}
\label{sec:experiments}

\subsection{Video-based person re-identification}
\label{sec:experiments_reid}

In this section, we evaluate the performance of the proposed COSAM layer in the video-based person re-ID pipeline by plugging it into two state-of-the-art deep architectures: ResNet50 (\cite{he2016deep}) \& SE-ResNet50 (\cite{hu2018squeeze}).
%on three video Re-ID datasets MARS\cite{zheng2016mars}, DukeMTMC-VideoReID\cite{wu2018exploit} \& iLIDS-VID\cite{wang2014person}. 

\subsubsection{Datasets and Evaluation protocol}

We evaluate the proposed COSAM module based Re-ID model on three commonly used video-based person Re-ID datasets: MARS (\cite{zheng2016mars}), DukeMTMC-VideoReID (\cite{wu2018exploit}) and iLIDS-VID (\cite{wang2014person}). The MARS dataset (\cite{zheng2016mars}) is the largest sequence-based person Re-ID dataset with 1261 identities and 20,478 video sequences, with multiple frames per person captured across 6 non-overlapping camera views. Among the total identities, 625 identities are used for training and the rest are used for testing. Additionally, 3,248 identities (disjoint with the train and test set) are used as distractors. DukeMTMC-VideoReID  (\cite{wu2018exploit})  is a subset of the DukeMTMC multi-camera dataset (\cite{ristani2016performance}), which was collected on outdoor scenario with varying viewpoint, illuminations, background and occlusions using 8 synchronized cameras. It contains 702 identities, each for training \& testing, and 408 identities as the distractors. There are 369,656 tracklets for training, and 445,764 frames for testing \& distractors. iLIDS-LID (\cite{wang2014person}) is a small dataset containing 600 sequences of 300 persons from two non-overlapping camera views. The sequences vary in length between 23 and 192 frames. As per the protocol followed in \cite{wang2014person,li2018diversity}, 10 random probe-gallery splits are used to perform experiments. 

We use the standard evaluation metrics as followed in the literature (\cite{zheng2016mars,li2018diversity,liu2018video,suh2018part}) viz., 1) \textit{Cumulative Matching Characteristics (CMC)} \& 2) \textit{Mean average precision (mAP)}. \textit{CMC} is based on the retrieval capability of the algorithm to find the correct identity within the top-k ranked matches. \textit{CMC} is used when only one gallery instance exists for every identity. We report rank-1, rank-5 and rank-20 CMC accuracies. The \textit{mAP} metric is used to evaluate algorithms in multi-shot re-identification settings where multiple instances of same identities are present in the gallery. 
%Hence, mAP is considered to be a suitable measure for both single-shot and multi-shot setting.

% It shows how often, on average, the correct person ID is included in the best K matches against the training set for each test image.

\vspace{-0.1cm}
\subsubsection{Implementation details}

The proposed method is implemented using the PyTorch framework (\cite{paszke2017automatic}) and is available online\footnote{\fontsize{7}{8}\selectfont\url{https://github.com/InnovArul/vidreid_cosegmentation}}. During training, every video consists of $N=4$ frames (as in baseline \cite{gao2018revisit}) and each frame is of height = $256$ and width = $128$. The images are normalized using the RGB mean and standard deviation of ImageNet (\cite{imagenet_cvpr09}) before passing to the network. The network is trained using the \textit{Adam} optimizer with the following hyper-parameters : number of reference feature maps $K = 3$, $\beta_1$ = 0.9, $\beta_2$ = 0.999, batch size = 32, initial learning rate = 0.0001, trade-off parameter between losses $\lambda = 1$ and COSAM dimension reduction size $D_R = 256$. We train the network for $\sim$ 60K iterations and the learning rate is multiplied by 0.1 after every 15K iterations. The implementation was done
on a machine with NVIDIA GeForce GTX 1080 Ti GPUs and takes around 8 hours to train a model with one GPU.

\subsubsection{Results \& Discussion \label{sec:results}}

\begin{figure*}[!htb]
\vspace{-0.3cm}
\begin{center}
\begin{subfigure}[b]{0.22\linewidth}
\begin{center}
\includegraphics[height=50px,width=\linewidth]{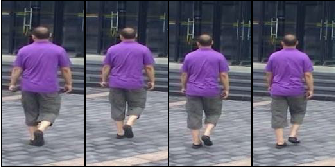}
\end{center}
\end{subfigure}
\quad
\begin{subfigure}[b]{0.22\linewidth}
\begin{center}
\includegraphics[height=50px,width=\linewidth]{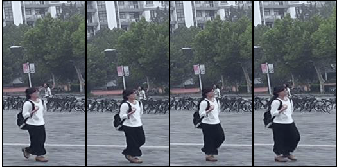}
\end{center}
\end{subfigure}
\quad
\begin{subfigure}[b]{0.22\linewidth}
\begin{center}
\includegraphics[height=50px,width=\linewidth]{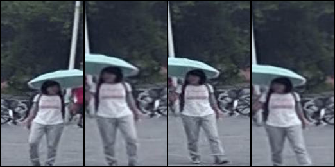}
\end{center}
\end{subfigure}
\quad
\begin{subfigure}[b]{0.22\linewidth}
\begin{center}
\includegraphics[height=50px,width=\linewidth]{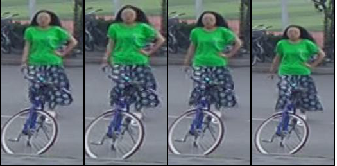}
\end{center}
\end{subfigure}
\begin{subfigure}[b]{0.22\linewidth}
\begin{center}
\includegraphics[height=50px,width=\linewidth]{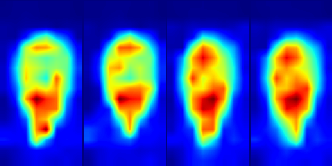}
\end{center}
\vspace*{-0.5cm}
\caption{}
\end{subfigure}
\quad
\begin{subfigure}[b]{0.22\linewidth}
\begin{center}
\includegraphics[height=50px,width=\linewidth]{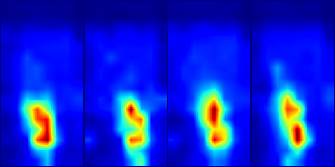}
\end{center}
\vspace*{-0.5cm}
\caption{}
\end{subfigure}
\quad
\begin{subfigure}[b]{0.22\linewidth}
\begin{center}
\includegraphics[height=50px,width=\linewidth]{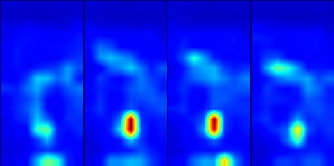}
\end{center}
\vspace*{-0.5cm}
\caption{}
\end{subfigure}
\quad
\begin{subfigure}[b]{0.22\linewidth}
\begin{center}
\includegraphics[height=50px,width=\linewidth]{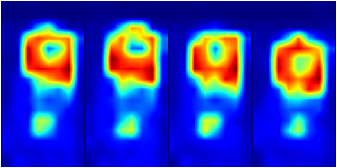}
\end{center}
\vspace*{-0.5cm}
\caption{}
\end{subfigure}
\vspace*{-0.3cm}
\caption{Visualization of COSAM's spatial attention mask in the Video-based person re-ID task. The second row shows the segmentation maps corresponding to the images in the first row.%\textcolor{magenta}{OCCLUSION, REARRANGE}
\label{fig:visualization}
}
\end{center}
\vspace*{-0.8cm}
\end{figure*}

In our experiments, every video of the person is split into multiple non-overlapping video-snippets of length $N$ frames and each snippet is passed through the network to obtain a snippet-level descriptor. Further, the video-snippet level descriptors are averaged to get the video-level descriptor. Then, these video-level descriptors are compared using the \textit{L$_2$} distance to calculate the CMC and mAP performances.

First, we present the results from our baseline method (\cite{gao2018revisit}) that re-visited the effect of various temporal aggregation layers with ResNet50 (\cite{he2016deep}) as the feature extractor. We extend this study by including yet another state-of-the-art architecture SE-ResNet50 (\cite{hu2018squeeze}\footnote{Winner of ILSVRC 2017 Image Classification Challenge\cite{imagenet_cvpr09}})
and present the quantitative results in Table \ref{tab:baseanalysis}. 

\begin{table}[!ht]
\vspace*{-0.2cm}
\scriptsize
\renewcommand{\arraystretch}{1.35}
\setlength\tabcolsep{2.5pt}
\begin{center}
\begin{tabular}{|l|c|c|c|c|c|c|c|c|c|c|}
\hline
\multirow{2}{1.5cm}{Feature extractor} & \multirow{2}{*}{\makecell[l]{Temp.\\ Agg}} &  \multicolumn{4}{c|}{MARS} & \multicolumn{4}{c|}{DukeMTMC-VideoReID} \\\cline{3-10}
 & & mAP & R1 & R5 & R20 & mAP & R1 & R5 & R20   \\\hline
\makecell[l]{ResNet50\\\cite{gao2018revisit}} & TP$_{avg}$ &   75.8 & 83.1 & 92.8 & 96.8 & 92.9 & 93.6 & \textbf{99.0} & \textbf{99.7}\\\hline
\makecell[l]{ResNet50\\\cite{gao2018revisit}} & TA & 76.7 & 83.3 & 93.8& \textbf{97.4} &93.2&93.9&98.9 &99.5\\\hline
\makecell[l]{ResNet50\\\cite{gao2018revisit}} & RNN & 73.8 & 81.6 & 92.8 & 96.7 &88.1 &88.7 &97.6& 99.3\\\hline
SE-ResNet50 & TP$_{avg}$ &  \textbf{78.1} & 84.0 & \textbf{95.2}& 97.1 &\textbf{93.5 }&93.7&\textbf{99.0} & \textbf{99.7}\\\hline
SE-ResNet50 & TA & 77.7 & \textbf{84.2} &94.7& \textbf{97.4}&93.1&\textbf{94.2}&\textbf{99.0} &\textbf{99.7}\\\hline
SE-ResNet50 & RNN &75.7& 83.1& 93.6& 96.0&92.4&94.0&98.4& 99.1\\\hline
% PartAlign\cite{suh2018part} & - & 72.2 & 83.0 & 92.8 &&&  \\\hline
\end{tabular}
\end{center}
\vspace*{-0.5cm}
\caption{Evaluation of a simple video-based Re-ID framework (as shown in Fig. \ref{fig:video-reid-pipeline}) on different feature extractor networks, feature aggregation techniques and datasets.\label{tab:baseanalysis} Best results are shown in \textbf{Bold}.}
\vspace*{-0.3cm}
\end{table}

\begin{table}[!htbp]
\vspace{-0.2cm}
\scriptsize
\renewcommand{\arraystretch}{1.2}
\setlength\tabcolsep{3.7pt}
\begin{center}
\begin{tabular}{|c|l|c|c|c|c|c|c|c|c|}
\hline
% \scriptsize
& \multirow{2}{*}{COSAM$_i$} & \multicolumn{4}{c|}{MARS} & \multicolumn{4}{c|}{DukeMTMC-VideoReID} \\\cline{3-10}
 & & mAP & R1 & R5 & R20 & mAP & R1 & R5 & R20  \\\hline
\parbox[t]{2mm}{\multirow{3}{*}{\rotatebox[origin=l]{90}{ResNet50\quad\quad}}} & \makecell[l]{No COSAM \\(\cite{gao2018revisit})} & 75.8 & 83.1 & 92.8 & 96.8&92.9&93.6&99.0  &99.7\\\cline{2-10}
 & COSAM$_2$ & 68.3 &77.7&90.1&96.1& 88.9 &90.2&98.4&99.0 \\\cline{2-10}
 & COSAM$_3$ & 76.9 & 82.7 & \textbf{94.3}& 97.3 &93.6&94.0&98.7& \textbf{99.9}\\\cline{2-10}
 & COSAM$_4$ & 76.8 & 82.9 & 94.2& 97.1 &93.8&\textbf{94.7}&98.7&99.7 \\\cline{2-10}
 & COSAM$_5$ & 76.6 & 82.8 & 93.9& 97.2 &93.2 &93.7&98.4&\textbf{99.9} \\\cline{2-10}
  & COSAM$_{3,4}$ & 76.4 &83.4&93.9&97.1&93.7&94.4&99.1&99.4\\\cline{2-10}
 & COSAM$_{3,5}$ & 76.9 &\textbf{83.7}&94.0&97.3&93.0&93.7&99.0&99.7\\\cline{2-10}
 & COSAM$_{4,5}$ & \textbf{77.2} & \textbf{83.7} & 94.1 &  \textbf{97.5} &\textbf{94.0}&94.4&\textbf{99.1}& \textbf{99.9}\\\cline{2-10}
  & COSAM$_{3,4,5}$ &76.6&83.2&93.7&97.3&93.1&93.6&98.7&99.4\\\hline\hline\hline
& No COSAM & 78.3 & 84.0 & 95.2& 97.1 & 93.5 & 93.7 & 99.0&99.7 \\\cline{2-10}
\parbox[t]{2mm}{\multirow{3}{*}{\rotatebox[origin=c]{90}{SE-ResNet50\quad\quad}}} & COSAM$_2$ & 67.0 & 77.9 & 90.4& 94.9 &92.2&94.0&98.9& 99.7\\\cline{2-10}
 & COSAM$_3$ &79.5 &{85.0}&94.7& 97.8&93.6&94.7&99.0&\textbf{99.9} \\\cline{2-10}
 & COSAM$_4$ & 79.8 &84.9&95.4&97.8& 94.0&\textbf{95.4}&99.0&\textbf{99.9} \\\cline{2-10}
 & COSAM$_5$ & \textbf{79.9} &84.5&\textbf{95.7}&97.9 & 93.9 &94.9&99.1& \textbf{99.9}\\\cline{2-10}
 & COSAM$_{3,4}$ &79.5 &84.8&94.7&97.6&93.7&94.7&98.7&99.7\\\cline{2-10}
 & COSAM$_{3,5}$ & 79.8 &\textbf{85.2}&95.5&98.0&93.9&94.2&99.3&99.9\\\cline{2-10}
 & COSAM$_{4,5}$ & \textbf{79.9} &{84.9}&95.5&\textbf{97.9}& \textbf{94.1}&\textbf{95.4}&\textbf{99.3}& 99.8 \\\cline{2-10}
 & COSAM$_{3,4,5}$ &\textbf{80.5}&\textbf{85.2}&95.5&\textbf{98.0}&\textbf{94.1}&\textbf{95.4}&\textbf{99.3}&\textbf{99.9}\\\hline
\end{tabular}
\end{center}
\vspace{-0.5cm}
\caption{Evaluation of the backbone feature extractors with COSAM and temporal aggregation layer as TP$_{avg}$. Here, \textit{COSAM$_i$} = plugging in COSAM layer after \textit{$i^{th}$ CNN} block.\label{tab:ablation_at_cnn_levels}}
\vspace{-0.3cm}
\end{table}

Based on our experiments in Table \ref{tab:baseanalysis}, we postulate certain key observations: {\textit{First}}, the selection of the backbone network can influence the holistic system performance. This is quite noteworthy since not much research on the influence of the backbone network on Re-ID performance has been conducted. \textit{Second}, it is observed that even a simple TP$_{avg}$ layer performs on-par with complex attention/RNN based aggregation layers, as also reported in \cite{chen2017cnn}. 

\paragraph{Location of the COSAM layer within the network} Next, to evaluate the effect of the COSAM module, we experiment by plugging-in the COSAM module after each \textit{CNN block} of the feature extractors and TP$_{avg}$ is used as the feature aggregation layer. The network is trained and evaluated on the MARS \& DukeMTMC-VideoReID datasets and the quantitative results are shown in Table \ref{tab:ablation_at_cnn_levels}. From the results, it can be inferred that the inclusion of the COSAM module improves the baseline network and it is effective in the deeper layers (COSAM$_3$, COSAM$_4$, COSAM$_5$), as the features in those layers are more discriminative and abstract than the features at shallow layer(s). We also experiment with the inclusion of multiple COSAM blocks simultaneously. It is found that COSAM$_{4,5}$ (plugging in COSAM$_{4}$ \& COSAM$_{5}$) achieves the best accuracy \textit{vs.} computation trade-off and is treated as our \textit{default} proposed architecture in the rest of the experiments.

\begin{table}[!htbp]
\vspace{-0.4cm}
\renewcommand{\arraystretch}{1.35}
\setlength\tabcolsep{2.2pt}
\scriptsize
\begin{center}
\begin{tabular}{|c|c|c|c|c|c|c|c|c|c|c|}
\hline
  & \multirow{2}{*}{\makecell{Temp. \\Agg.}} & \multirow{2}{*}{COSAM$_i$} & \multicolumn{3}{c|}{MARS} & \multicolumn{3}{c|}{Duke}  & \multicolumn{2}{c|}{iLIDS-VID} \\\cline{4-11}
& & & mAP & R1 & R5 & mAP & R1 & R5 & R1 & R5 \\\hlineB{3}
% AlexNet & 2012 & 2.47M &&&&&&&&&&&& \\\hline
% VGGNet16  & 2013 &&&&&&&&&&&&& \\\hline
% Inception-v4  & 2015 &&&&&&&&&&&&& \\\hline
\parbox[t]{2mm}{\multirow{6}{*}{\rotatebox[origin=l]{90}{ResNet50\quad\quad}}} &  \makecell[c]{TP$_{avg}$\\\cite{gao2018revisit}} & - & 75.8 & 83.1 & 92.8 & 92.9 & 93.6 & 99.0 & 73.9 & 92.6\\\cline{2-11}
& TP$_{avg}$ & COSAM$_{4,5}$  & \textcolor{red}{\textbf{77.2}} & \textcolor{blue}{\textbf{83.7}} & 94.1 & \textcolor{red}{\textbf{94.0}} & 94.4 &99.1 & \textcolor{blue}{\textbf{75.5}} & 94.1 \\\clineB{2-11}{3}
 & \makecell[c]{TA\\\cite{gao2018revisit}} & - &  76.7 & 83.3 & 93.8 &  93.2&93.9&98.9 & 72.3 & 92.4 \\\cline{2-11}
  & TA  & COSAM$_{4,5}$ & 76.9 & 83.6& 93.7 &93.4&\textcolor{blue}{\textbf{94.6}}&98.9 &  74.9 & 94.4 \\\clineB{2-11}{3}
 & \makecell[c]{RNN\\\cite{gao2018revisit}} & - &  73.8 & 81.6 & 92.8 &   88.1 &88.7 &97.6 &  68.5 & 93.2\\\cline{2-11}
  & RNN & COSAM$_{4,5}$ & 74.8 & 82.4 & 93.9 &90.4&91.7&98.3 & 68.9 & 93.1\\\hlineB{3}
% \Xhline{2\arrayrulewidth}
\addlinespace[0.1cm]\hlineB{3}
\parbox[t]{2mm}{\multirow{6}{*}{\rotatebox[origin=l]{90}{SE-ResNet50}}} &  TP$_{avg}$ & -&  78.1 & 84.0 & 95.2 & 93.5&93.7& 99.0 & 76.9 & 93.9  \\\cline{2-11}
 & TP$_{avg}$ & COSAM$_{4,5}$ & \textcolor{red}{\textbf{79.9}} & 84.9& 95.5 &\textcolor{red}{\textbf{94.1}}&\textcolor{blue}{\textbf{95.4}}&99.3 & \textcolor{blue}{\textbf{79.6}} & 95.3 \\\clineB{2-11}{3}
 & TA & - & 77.7 & 84.2 & 94.7 &93.1&94.2&99.0 & 74.7 & 93.2 \\\cline{2-11}
& TA & COSAM$_{4,5}$ & 79.1 &\textcolor{blue}{\textbf{85.0}} & 94.9  &  94.1 & \textcolor{blue}{\textbf{95.3}}& 98.9 & 77.1 & 94.7 \\\clineB{2-11}{3}
 &  RNN & - & 75.7 & 83.1 & 93.6  & 92.4&94.0&98.4 & 77.4 & 94.4 \\\cline{2-11}
  & RNN  & COSAM$_{4,5}$ & 76.0 & 83.4 & 93.9 & 92.5&93.9&98.3 & 77.8 & 97.3\\\hlineB{3}
\end{tabular}
\end{center}
\vspace{-0.5cm}
\caption{Comparison of the baseline models with best performing COSAM-configuration (COSAM$_{4,5}$) along with different feature extractor networks, feature aggregation techniques and datasets. Here, COSAM$_{4,5}$ = COSAM layer is placed after 4$^{th}$ and 5$^{th}$ CNN blocks of the baseline model, Duke = DukeMTMC-VideoReID dataset. Best mAP \& CMC Rank-1 per backbone network are shown in \textcolor{red}{\textbf{red}} and \textcolor{blue}{\textbf{blue}} colors respectively. mAP is not applicable for iLIDS-VID due to single gallery instance per probe.
\label{tab:baseline_compare}}
\vspace{-0.5cm}
\end{table}

\paragraph{Visualizations} To demonstrate the interpretability of our proposed method, we visualize the spatial attention masks of the COSAM$_4$ layer in the SE-ResNet50+COSAM$_{4,5}$ model trained on the MARS dataset (Fig.~\ref{fig:visualization}). The frames exhibit varying conditions such as scale, pose, viewpoint changes and partial occlusions. In Fig.~\ref{fig:visualization}(a), the predicted attention mask is able to focus on the person and avoid background features. In Fig.~\ref{fig:visualization}(b), despite the person occupying comparatively a small region of the frame, the COSAM layer still successfully focuses on the person based on task-relevant consensus. Although the buildings and trees are common in all the frames, our Co-segmentation inspired architecture specifically trained for Re-ID ignores the background regions.
In Fig.~\ref{fig:visualization}(c), it can be observed that the spatial attention identifies the accessory (umbrella) carried by the person. Identifying the person with the aid of their belongings is one of the significant ways to discriminate the person by appearance. In Fig.~\ref{fig:visualization}(d), the partial occlusion scenario is handled successfully by avoiding the occluding object (cycle). 

\begin{table}[!ht]
\scriptsize
\vspace{-0.3cm}
\renewcommand{\arraystretch}{1.2}
\setlength\tabcolsep{3pt}
\begin{center}
\begin{tabular}{|l|p{0.7cm}|c|c|c|c|}
\hline
\multirow{2}{*}{Network} & \multirow{2}{0.7cm}{Deep model?} &  \multicolumn{4}{c|}{\textbf{MARS}} \\\cline{3-6}
 & & mAP & R1 & R5 & R20   \\\hline
LOMO+XQDA (\cite{liao2015person}) & No & 16.4 & 30.7 & 46.6 & 60.9 \\\hline
JST-RNN (\cite{zhou2017see}) & Yes & 50.7 & 70.6 & 90.0 & 97.6  \\\hline
QAN (\cite{liu2017quality}) & Yes & 51.7 & 73.7 & 84.9 & 91.6 \\\hline
Context Aware Parts (\cite{li2017learning}) & Yes & 56.1 & 71.8 & 86.6 & 93.0 \\\hline
IDE+XQDA+ReRanking (\cite{zhong2017re})& Yes & 68.5 & 73.9 & - & -\\\hline
TriNet (\cite{hermans2017defense})& Yes & 67.7 & 79.8 & 91.4 & -\\\hline
Region QEN (\cite{song2018region})& Yes & 71.1 & 77.8 & 88.8 & 94.1 \\\hline
Comp. Snippet Sim.(\cite{chen2018video}) & Yes& 69.4 & 81.2 & 92.1 & -  \\\hline
Part-Aligned (\cite{suh2018part}) & Yes & 72.2 & 83.0 & 92.8 & 96.8  \\\hline
RevisitTempPool (\cite{gao2018revisit})& Yes  &76.7&83.3&93.8&97.4 \\\hlineB{2}
(\cite{gao2018revisit}) + SE-ResNet50 + TP$_{avg}$ & Yes  &78.1 & 84.0 & 95.2 & 97.1 \\\hlineB{2}
\makecell[l]{SE-ResNet50 + COSAM$_{4,5}$\\ + TP$_{avg}$(ours)} & Yes  & \textbf{79.9} & \textbf{84.9}& \textbf{95.5} & \textbf{97.9} \\\hline
\makecell[l]{SE-ResNet50 + COSAM$_{4,5}$\\ + TP$_{avg}$(ours) + Re-ranking (\cite{zhong2017re})} & Yes  & \textbf{87.4} & \textbf{86.9} & \textbf{95.5} & \textbf{98.0}
\\\hline
% \addlinespace[0.1cm]
\hline
\multirow{2}{*}{Network} & \multirow{2}{0.7cm}{Deep model?} &  \multicolumn{4}{c|}{\textbf{DukeMTMC-VideoReID}} \\\cline{3-6}
 & & mAP & R1 & R5 & R20   \\\hline
ETAP-Net (\cite{suh2018part}) & Yes & 78.34 & 83.62 & 94.59 & 97.58 \\\hline
% \makecell[l]{ResNet50 + COSAM$_{4,5}$ + \\TP$_{avg}$(ours)} & Yes  &  & &  & \\\hline
RevisitTempPool (\cite{gao2018revisit}) & Yes & 93.2 & 93.9 & 98.9 & 99.5\\\hlineB{2}
(\cite{gao2018revisit}) + SE-ResNet50 + TP$_{avg}$ & Yes  & 93.5 & 93.7 & 99.0 & 99.7 \\\hlineB{2}
\makecell[l]{SE-ResNet50 + COSAM$_{4,5}$ + \\TP$_{avg}$(ours)} & Yes  & \textbf{94.1} & \textbf{95.4} & \textbf{99.3}  &  \textbf{99.8}
\\\hline
\end{tabular}
\end{center}
\vspace{-0.6cm}
\caption{Comparison of our best model with state-of-the-art methods on MARS \& DukeMTMC-VideoReID datasets. \label{tab:SOTA_compare}}
\vspace{-0.3cm}
\end{table}

\begin{table}[!htbp]
\vspace{-0.1cm}
\footnotesize
\renewcommand{\arraystretch}{1.2}
\setlength\tabcolsep{4pt}
\begin{center}
\begin{tabular}{|l|c|c|c|}
\hline
\multirow{2}{*}{Method} & \multicolumn{3}{c|}{iLIDS-VID}  \\\cline{2-4}
 & R1 & R5 & R20 \\\hline
Top push video Re-ID (\cite{you2016top}) & 56.3 &87.6&98.3 \\\hline
JST-RNN (\cite{zhou2017see}) & 55.2 & 86.5  & 97.0\\\hline
Joint ST pooling (\cite{xu2017jointly}) & 62.0 & 86.0 & 98.0\\\hline
Region QEN (\cite{song2018region}) & 77.1 & 93.2 & 99.4 \\\hline
RevisitTempPool (\cite{gao2018revisit}) & 73.9 & 92.6 & 98.41 \\\hline
\makecell[l]{\cite{gao2018revisit} + SE-ResNet50 + TP$_{avg}$} & 76.87 & 93.94 & 99.07 \\\hline
\makecell[l]{SE-ResNet50 + COSAM$_{4,5}$ + TP$_{avg}$(ours)} & \textbf{79.61} & \textbf{95.32} & \textbf{99.8} \\\hline
\end{tabular}
\end{center}
\vspace{-0.6cm}
\caption{Comparison of the state-of-the-arts in iLIDS-VID re-ID dataset. \label{tab:SOTA_ilidsvid}}
\vspace{-0.6cm}
\end{table}

\paragraph{Effect of COSAM in the baseline model}
To understand the significance of the COSAM layer, we incorporate our best performing COSAM$_{4,5}$ into baseline video-based Re-ID pipelines with two feature extractors (ResNet50 and SE-ResNet50) and three different temporal aggregation layers (TP$_{avg}$, TA, RNN) (\cite{gao2018revisit}). Table \ref{tab:baseline_compare} represents the performance evaluation of the models. Our COSAM-based networks show consistent performance improvement (both CMC Rank and mAP) over the baseline models in all three datasets. Between the backbone networks, SE-ResNet50 outperforms ResNet50 in both baselines and proposed case studies, highlighting the importance of a better backbone network selection. Among the temporal aggregation modules, although more or less similar performance is exhibited by 
TP$_{avg}$, TA and RNN, the former (TP$_{avg}$) results in the best mAP 
values in both MARS and DukeMTMC-VideoReID datasets \& best CMC Rank-1 in iLIDS-VID. In particular, COSAM improves the mAP by 1.4\% (ResNet50) \& 1.8\% (SE-ResNet50) in MARS and 1.1\%(ResNet50) \& 0.6\% (SE-ResNet50) in 
DukeMTMC-VideoReID respectively. 
%Regarding CMC Rank, both the TP$_{avg}$ and TA perform equally well resulting in the best ranks, 
Regarding the CMC Rank, we observe an improvement of 0.6\% (ResNet50) \& 0.9\% (SE-ResNet50) in MARS, 0.8\% (ResNet50) \& 1.7\% (SE-ResNet50) in DukeMTMC-VideoReID and 1.6\% (ResNet50) \& 2.7\% (SE-ResNet50) in iLIDS-VID.

\paragraph{Comparison with state-of-the-art methods}
We compare our method with the state-of-the-arts \cite{liao2015person,hermans2017defense,liu2017quality,li2017learning,zhong2017re,hermans2017defense,song2018region,chen2018video,suh2018part, gao2018revisit} in MARS and DukeMTMC-VideoReID datasets and the results are shown in Table \ref{tab:SOTA_compare}. It is observed that our proposed COSAM module applied into SE-ResNet50 (COSAM$_{4,5}$) along with TP$_{avg}$ achieves the best performance. In particular, our approach has $\sim$0.9\% improvement in CMC Rank-1 as well as 1.8\% improvement in mAP in the MARS dataset over the best performing method (\cite{gao2018revisit} + SE-ResNet50 + TP$_{avg}$). Apart from this, applying re-ranking (\cite{zhong2017re}) further increases the performance to +2.0\% CMC Rank-1 and +7.5\% mAP. Intuitively, such an improved CMC Rank-1 (86.9\%) shows that majority of the subjects are correctly identified in the first rank, whereas the improved mAP result (87.4\%) denotes that multiple instances of the person are ranked precisely at the top in a multi-shot setting that is significant in retrieval problems. We attribute this improvement to the effectiveness of the COSAM layer in suppressing noise and aiding the network learn about identifying relevant common objects. Similarly, our COSAM layer with SE-ResNet50 achieves 0.6\% improvement in mAP and 1.7\% improvement in CMC Rank-1 with DukeMTMC-VideoReID dataset. The performance comparison of iLIDS-VID dataset in the Table \ref{tab:SOTA_ilidsvid} shows that COSAM layer with SE-ResNet50 improves +2.8\% in CMC Rank-1 over the baseline method (\cite{gao2018revisit}).

\subsubsection{Ablation studies\label{sec:ablation_reid}}

\paragraph{Effect of different frame lengths ($N$): } We study the effect of the number of frames in a video on the performance of our best performing model. In particular, we analyze with frame lengths of $N$ = 2, 4 and 8 in  SE-ResNet50+COSAM$_{4,5}$+TP$_{avg}$ and the results are shown in Table~\ref{tab:ablation_Ns}. We found $N=4$ frames to be optimal similar to \cite{gao2018revisit}. 

\begin{table}[!ht]
\vspace{-0.2cm}
\footnotesize
\renewcommand{\arraystretch}{1.2}
\setlength\tabcolsep{4pt}
\begin{center}
\begin{tabular}{|c|c|c|c|c|c|c|c|c|}
\hline
\multirow{2}{*}{frame length} & \multicolumn{4}{c|}{MARS} & \multicolumn{4}{c|}{DukeMTMC-VideoReID} \\\cline{2-9}
 & mAP & R1 & R5 & R20 & mAP & R1 & R5 & R20  \\\hline
 $N = 2$ & 78.1& 83.5&94.3&98.1& 94.0&94.3&99.1&\textbf{99.9}\\\hline
 $N = 4$ & \textbf{79.9} &\textbf{84.9} &\textbf{95.5} & \textbf{97.9} &\textbf{94.1}&\textbf{95.4}&\textbf{99.3} & 99.8\\\hline
 $N = 8$ & 77.4 & 84.6 & 94.2 &  97.0 & 92.1&91.9&99.0&99.6\\\hline
\end{tabular}
\end{center}
\vspace{-0.6cm}
\caption{Evaluation of the influence of track length $T$ on Re-ID performance of the best performing model \textit{SE-ResNet50+COSAM$_{4,5}$+TP$_{avg}$}. \label{tab:ablation_Ns}}
\vspace{-0.7cm}
\end{table}

\paragraph{Attribute-wise performance gains:} To understand the importance of COSAM in capturing attributes, we conduct attribute-wise empirical studies on the DukeMTMC-VideoReID dataset and present the results in Table \ref{tab:attribute-duke}. The significant improvements on attributes such as handbag, hat and backpack show that COSAM is indeed capturing the person's attributes.

\begin{table}[!ht]
\vspace{-0.2cm}
\centering
\scriptsize
\renewcommand{\arraystretch}{1.0}
\setlength\tabcolsep{2pt}
 \begin{tabular}{|p{1.6cm}|p{0.6cm}|p{0.6cm}|p{0.6cm}|p{0.6cm}|p{0.6cm}|p{0.6cm}|p{0.6cm}|p{0.6cm}|p{0.6cm}|} 
 \hline
\multirow{2}{*}{Model} & \multicolumn{3}{c|}{Handbag} & \multicolumn{3}{c|}{Hat} & \multicolumn{3}{c|}{Backpack} \\\cline{2-10}
& mAP & R1 & R5 & mAP & R1 & R5 & mAP & R1 & R5 \\\hline
R50+TP & 91.2 &92.0&\textbf{100.0}&91.1&91.7&97.5&92.8&93.9&98.6 \\\hline
R50+C$_{4,5}$+TP & \textbf{95.2}&\textbf{96.0}&\textbf{100.0}&\textbf{93.5}&\textbf{94.2}&\textbf{97.5}&\textbf{95.1}&\textbf{96.4}&\textbf{99.8} \\\hline\hline
SE50+TP &94.1&97.3&\textbf{100.0}&92.7&94.2&99.2&94.3&95.6&99.1 \\\hline
SE50+C$_{4,5}$+TP & \textbf{96.0}&\textbf{100.0}&\textbf{100.0}&\textbf{93.9}&\textbf{96.7}&\textbf{99.5}&\textbf{95.4}&\textbf{97.1}&\textbf{100.0}  \\\hline
\end{tabular}
\vspace{-0.3cm}
\caption{\small Attribute-wise perf. comparison on Duke reveals the effectiveness of COSAM to capture features of person's accessories. Here, R50=ResNet50, SE50=SE-ResNet50, C$_{4,5}$=COSAM$_{4,5}$, TP = Temporal average pooling. \label{tab:attribute-duke}}
\vspace{-0.5cm}
\end{table}

\paragraph{Selection of frames}
We perform experiments with two schemes of frame selection during training namely: 1) \textit{Sequential}: a continuous sequence of $N$ frames are selected, 2) \textit{Random}: random sampled $N$ frames from the whole video sequence. The quantitative results are shown in Table \ref{tab:frame_selection}. It is observed that the performance of training using randomly sampled frames is inferior to the performance of training using sequentially sampled frames. 

\begin{table}[!htbp]
\vspace{-0.2cm}
\footnotesize
\renewcommand{\arraystretch}{1.2}
\setlength\tabcolsep{4pt}
\begin{center}
\begin{tabular}{|c|c|c|c|c|c|c|c|c|}
\hline
\multirow{2}{*}{frame selection} & \multicolumn{4}{c|}{MARS} & \multicolumn{4}{c|}{DukeMTMC-VideoReID} \\\cline{2-9}
 & mAP & R1 & R5 & R20 & mAP & R1 & R5 & R20  \\\hline
 sequential & \textbf{79.9} &{\textbf{84.9}}&\textbf{95.5}&\textbf{97.9}& \textbf{94.1}&\textbf{95.4}&\textbf{99.3}& \textbf{99.8} \\\hline
 random &77.5 & 83.3&93.6&97.0 &90.2&90.6&98.3 &99.6 \\\hline
\end{tabular}
\end{center}
\vspace{-0.6cm}
\caption{Evaluation of the influence of frame selection on Re-ID performance of the best performing model \textit{SE-ResNet50+COSAM$_{4,5}$+TP$_{avg}$}. \label{tab:frame_selection}}
\vspace{-0.5cm}
\end{table}

\paragraph{Cross-dataset test performance}
We analyze the cross-dataset performance of the best performing model
(\textit{SE-ResNet50+COSAM$_{4,5}$+ TP$_{avg}$}) against a base-model without using COSAM layer. The results are shown in Table \ref{tab:cross-dataset}. By training on MARS dataset and testing on DukeMTMC-VideoReID dataset, the former model outperforms the latter by 2.8\% mAP and 3.5\% CMC Rank-1. Similar performance of COSAM-based model is observed while training on DukeMTMC-VideoReID dataset and testing on MARS dataset (Improvement of 0.9\% in mAP and 0.7\% in CMC Rank-1).

\begin{table}[!htbp]
\footnotesize
\renewcommand{\arraystretch}{1.2}
\setlength\tabcolsep{4pt}
\begin{center}
\begin{tabular}{|c|c|c|c|c|c|c|}
\hline
& Train set & Test set &  mAP & R1 & R5 & R20   \\\hline
No COSAM & MARS & \makecell{DukeMTMC} &32.0 &33.3&53.3&67.1 \\\hline
COSAM$_{4,5}$ & MARS & \makecell{DukeMTMC} & \textbf{34.8}&\textbf{36.8}&\textbf{54.1}&\textbf{67.9} \\\hline
No COSAM & \makecell{DukeMTMC} & MARS & 25.0 & 41.7&54.4&65.3  \\\hline
COSAM$_{4,5}$ & \makecell{DukeMTMC} & MARS & \textbf{25.9}&\textbf{42.4}&\textbf{56.0}&\textbf{65.8} \\\hline
\end{tabular}
\end{center}
\vspace{-0.6cm}
\caption{Cross-dataset performance of the best performing model with \textit{SE-ResNet50} as the feature extractor and  \textit{TP$_{avg}$} as the temporal aggregation layer. \label{tab:cross-dataset}Here \textit{DukeMTMC} = DukeMTMC-VideoReID.}
\vspace{-0.4cm}
\end{table}

\paragraph{Spatial \textit{vs.} Channel attention }
To evaluate the effect of the attention steps in the COSAM layer, we investigate the individual attention steps (Spatial and Channel). The results in Table  \ref{tab:analysis_attention} shows that the inclusion of both spatial and channel attention steps together achieves superior performance.

\begin{table}[!htbp]
\footnotesize
\renewcommand{\arraystretch}{1.2}
\setlength\tabcolsep{4pt}
\begin{center}
\begin{tabular}{|c|c|c|c|c|c|c|c|c|}
\hline
\multirow{2}{*}{Attention layer} & \multicolumn{4}{c|}{MARS} & \multicolumn{4}{c|}{DukeMTMC-VideoReID} \\\cline{2-9}
 & mAP & R1 & R5 & R20 & mAP & R1 & R5 & R20  \\\hline
 Only spatial att. &78.8&84.1&94.9&97.7&93.6&93.9&99.0&\textbf{99.9} \\\hline
 Only Channel att. &79.0&84.3&95.0&97.8&93.8&94.4&99.1&99..7 \\\hline
  Both  & \textbf{79.9} &{\textbf{84.9}}&\textbf{95.5}&\textbf{97.9}& \textbf{94.1}&\textbf{95.4}&\textbf{99.3}& {99.8} \\\hline
\end{tabular}
\end{center}
\vspace{-0.6cm}
\caption{Evaluation of the influence of Co-segmentation based attention layers on Re-ID performance of the best performing model \textit{SE-ResNet50+COSAM$_{4,5}$+ TP$_{avg}$}. \label{tab:analysis_attention}}
\vspace{-0.6cm}
\end{table}

\paragraph{Comparison with Non-Local attention module (NLM)}  In this section, we compare and contrast the COSAM module with non-local attention modules (NLM) proposed by  \cite{wang2018non} for video-based tasks. Though both modules are conceptually similar, there are succinct differences between COSAM and NLM.

\begin{table}[!ht]
\centering
\footnotesize
\renewcommand{\arraystretch}{1.0}
\setlength\tabcolsep{2pt}
 \begin{tabular}{|l|c|c|c|} 
 \hline
 \multicolumn{2}{|c|}{Module}& \#Params & \#FLOPs \\ [0.5ex]
 \hline
 \parbox[t]{4mm}{\multirow{4}{*}{\rotatebox[origin=l]{90}{\textbf{NLM}}}} & Gauss. & 4.2M & 4.3B \\\cline{2-4}
&Gaussian embedding & 8.39M & 8.59G \\\cline{2-4}
&Concatenation & 8.4M & 8.72G\\\cline{2-4}
&Dot product &  8.39M & 8.59G \\\hline
\multicolumn{2}{|c|}{\textbf{COSAM (ours)}} & \textbf{1.6M} & \textbf{0.57G}\\\hline
\end{tabular}     
\vspace{-0.2cm}
\caption{COSAM \textit{vs.} NLM (\cite{wang2018non}) (input = {$4$$\times$$2048$$\times$$16$$\times$$8)$}. % ($N\times D\times H\times W$) 
Observation: COSAM uses $\sim4x$ less memory and $\sim16x$ less computation than NLM.
}\label{tab:cosam_vs_nlm}
\vspace{-0.4cm}
\end{table}

\begin{table}[!ht]
\centering
\footnotesize
 \begin{tabular}{|l|c|c|c|c|c|} 
 \hline
\multirow{2}{*}{Model} & \multirow{2}{*}{\#Params} & \multirow{2}{*}{\#FLOPs} & \multicolumn{3}{c|}{MARS} \\ [0.5ex] 
 \cline{4-6}
 & & & mAP & R1 & R5 \\
 \hline
R50+NLM$_{4,5}$+TP & 34.31M & 27.11B &76.9&83.2& \textbf{94.2} \\\hline
R50+COSAM$_{4,5}$+TP & 26.22M & 17.24B &\textbf{77.2}&\textbf{83.7}&94.1\\\hline\hline
SE50+NLM$_{4,5}$+TP & 36.85M & 26.74B & 77.9 & 83.3 & 94.7\\\hline
SE50+COSAM$_{4,5}$+TP & 28.76M & 16.86B & \textbf{79.9} & \textbf{84.9} & \textbf{95.5} \\\hline
\end{tabular}
\vspace{-0.2cm}
\caption{Comparison of COSAM \textit{vs.} NLM (\cite{wang2018non}) on MARS dataset. Here, R50 = ResNet50, SE50 = SE-ResNet50, NLM$_{4,5}$ = Non-local module after $4th$ and $5th$ block, COSAM$_{4,5}$ = COSAM module after $4th$ and $5th$ block.} \label{tab:cosam_vs_nlm_mars}
% }
\vspace{-0.4cm}
\end{table}

NLM is designed as a residual module that aggregates the features from neighboring / reference feature maps, where as COSAM activates common salient regions between feature maps and consists of a mandatory spatial and channel attention step to pass through only the common features. Both NLM and COSAM incurs $O(N * n^2)$ complexity for comparing descriptors of each frame with its reference feature maps. Here N = number of frames, $ n = H \times W$. However, NLM involves channel reduction step followed by feature comparison and aggregation, then channel expansion. On the other hand, COSAM only requires dimension reduction step followed by feature comparison which makes COSAM lightweight in terms of number of parameters and FLOPs. Table \ref{tab:cosam_vs_nlm} illustrates that COSAM uses $\sim4x$ less memory and $\sim16x$ less computation than NLM (\cite{wang2018non}) for an input of dimension $4 \times 2048 \times 16 \times 8$.
Further, we conduct experiments with Gaussian embedding instantiation of NLM on MARS dataset and show the results in Table \ref{tab:cosam_vs_nlm_mars}. Results from Table \ref{tab:cosam_vs_nlm_mars} reveals that COSAM performs superior to NLM in terms of mAP and rank-1 accuracy. Specifically, with SE-ResNet50 backbone, COSAM model (SE-ResNet50 + COSAM$_{4,5}$ + TP) performs +2.0\% and +1.6\% in mAP, rank-1 accuracy than the NLM model (SE-ResNet50 + NLM$_{4,5}$ + TP).

\begin{table*}
% \vspace{-0.4cm}
\begin{center}
\begin{tabular}{|l|c|c|c|c|c|c|c|c|c|c|}
\hline
\multirow{2}{*}{Methods}&\multirow{2}{*}{Year}&\multirow{2}{*}{Object detector}&\multicolumn{4}{|c|}{MSRVTT}&\multicolumn{4}{|c|}{MSVD} \\
\cline{4-11}
& & & B & M & R & C& B & M & R & C \\
\hline
POS+VCT (\cite{hou2019joint}) &ICCV-2019 &-   & 42.3 & \textbf{29.7} & \textbf{62.8} & 49.1  & 52.8 & 36.1 & 71.8 & 87.8\\
ORG-TRL (\cite{zhang2020object}) &CVPR-2020 & Faster RCNN & \textbf{43.6} & 28.8 & 62.1 & \textbf{50.9} & \textbf{54.3} & \textbf{36.4} & \textbf{73.9} & \textbf{95.2}\\
\hline\hline
OA-BTG (\cite{zhang2019object})&CVPR-2019 &Mask RCNN & \textbf{41.4} & 28.2 & -   & 46.9 & \textbf{56.9} & 36.2 & -   & 90.6 \\
STG-KD (\cite{pan2020spatio}) &CVPR-2020 &Faster RCNN & 40.5 & \textbf{28.3} & \textbf{60.9} & \textbf{47.1} & 52.2 & \textbf{36.9} &\textbf{73.9} & \textbf{93.0}\\
\hline\hline
SA-LSTM (\cite{wang2018reconstruction})&CVPR-2018 &- & 36.3 & 25.5 & 58.3 & 39.9 & 45.3 & 31.9 & 64.2 & 76.2\\
M3 (\cite{wang2018m3}) &CVPR-2018 &- & 38.1 & 26.6 & - & - & 52.8 & 33.3 & - & - \\
RecNet (\cite{wang2018reconstruction}) &CVPR-2018 &-   & 39.1 & 26.6 & 59.3 & 42.7 & \textbf{52.3} & 34.1 & 69.8 & 80.3\\
PickNet (\cite{chen2018less})&ECCV-2018 &-   & 41.3 & 27.7 & 59.8 & 44.1 &\textbf{52.3} & 33.3 & 69.6 & 76.5 \\
MARN (\cite{pei2019memory})&CVPR-2019 &-   & 40.4 & \textbf{28.1} & 60.7 & \textbf{47.1} & 48.6 & 35.1 & 71.9 & 92.2\\
Ours &- &-   & \textbf{41.4} & 27.8 & \textbf{61.0} & 46.5 & 50.7 & \textbf{35.3} & \textbf{72.1} & \textbf{97.8}\\
\hline
\end{tabular}
\end{center}
\vspace{-0.6cm}
\caption{Performance analysis of video captioning task on MSRVTT and MSVD datasets. First group of methods optimize on decoder. Second group of methods enhance encoder visual features by making use of object detectors. Third group of methods enhance encoder features without the usage of object detectors. Here, B = BLEU@4, M = METEOR, R = ROUGE\_L, C = CIDEr.\label{tab:sota}}
\vspace{-0.4cm}
\end{table*}

\subsection{Video captioning}
\label{sec:experiments_caption}

In this section, we describe the experiments and results of COSAM-based video captioning model along with the comparison of state-of-the-arts for this task.

\subsubsection{Datasets}
We use two challenging benchmark datasets: MSR-VTT (\cite{xu2016msr}) and MSVD (\cite{chen2011collecting}). 

\paragraph{MSR-VTT} It contains 10,000 video clips from 20 wide-range of categories (cooking, animations, sports, etc) with an average video length of 20 seconds. Each video has 20 human-annotated English captions. We follow the standard split followed in \cite{pan2016jointly,pei2019memory,zhang2019object} for train, validation and test as follows: 6513 video clips for training set, 497 for cross-validation set and remaining 2990 for the test set.

\paragraph{MSVD} It contains total of 1970 videos with approx. 40 English captions per video. We follow standard protocol split (\cite{pan2016jointly,pei2019memory,zhang2019object}) of dividing 1970 videos into train = 1200, validation = 100, and test = 670 videos for performing experiments. 

\subsubsection{Evaluation metrics}

We use the following metrics found in the literature to quantitatively evaluate our models performance: BLEU@4 (\cite{papineni2002bleu}), METEOR (\cite{denkowski2014meteor}), ROUGE\_L (\cite{lin2004rouge}) and CIDEr (\cite{vedantam2015cider}). 
BLEU@N matches n-grams between the generated captions and the ground truth, while METEOR metric is based on word-to-word matching between the generated captions and the ground truth, ROUGE\_L is Longest common subsequence based metric (LCS) and CIDEr calculates n-gram similarity between the generated caption and the ground truth. The above mentioned metrics are computed using the Microsoft COCO evaluation server (\cite{chen2015microsoft}).

\subsubsection{Implementation Details}

We implement our model using the PyTorch (\cite{paszke2019pytorch}) deep learning framework. 
During training, for every video, we use uniformly subsampled $T=10$ frames as input with correct temporal order. 
The frames are resized to a spatial resolution of $256 \times 256$ and are center cropped to $224 \times 224$.
% The frames are resized to spatial resolution of $224\times224$. We use standard data augmentations techniques of random horizontal translation, affine transformation and color jitter.
We use the 2D CNN ResNet-101 pre-trained on ImageNet (\cite{deng2009imagenet}) dataset to extract 2D features in GSB. Also, we extract intermediate features after the $5^{th}$ block (dimension $T \times {1024\times14\times14}$) as input features to the CoSB branch. For acquiring 3D features in GSB, we use 3D ResNeXt-101 pre-trained on Kinetics dataset. For every frame, a clip of 16 adjacent frames is used as input for ResNeXt-101 to get per-frame 3D feature.
The following hyper-parameters are used in network design: $C_{R} = 512, N_o = 5$. 

%The 2D and 3D features in GSB branch are projected to 512 dimensions each before getting concatenated.

For the transformer decoder, we adopt \cite{chen2018tvt} as our transformer architecture. We build a vocabulary of size $11$K and $6.5$K for MSR-VTT and MSVD respectively. During this process, punctuations are removed from every sentence and a word is removed if its count is one or less in whole dataset to avoid imbalanced distributions. During training, the maximum length of the sentence is set to 20. We use 2 multi-head self attention layers (one to encode visual features and the other to generate captions) in the decoder. We greedily output words with beam size of 1. The self-attention module has 8 attention heads and an MLP to output features with $1024$ dimension. 

We use a cross validation set to tune the hyper-parameters. The hyper-parameters are set as follows: batch size = $64$, learning rate = $10^{-4}$, the learning rate is decayed by multiplying $0.8$ after every $200$ epochs, weight decay = $5\times 10^{-4}$, dropout probability = $0.3$. The model is optimized using Adam optimizer with default parameters ($\beta_{1}=0.9$, $\beta_{2}=0.999$) for $650$ epochs with early stopping to get the best performing model.  For MSR-VTT dataset, we set $\lambda = 1$ and for MSVD dataset, $\lambda = 2$. $\lambda_{KL}$ is set to $4$ for both datasets.

\begin{figure*}[!ht]
\vspace{-0.2cm}
\begin{subfigure}{0.33\textwidth}
\begin{center}
% \fbox{\rule{0pt}{2in} \rule{.9\linewidth}{0pt}}
\includegraphics[height=0.4\linewidth,width=0.9\linewidth]{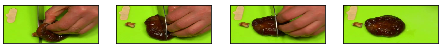}
\includegraphics[height=0.4\linewidth,width=0.9\linewidth]{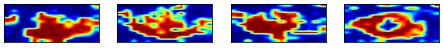}
\vspace{-0.2cm}
\caption{
\label{fig:qual0}
\tabular[t]{@{}l@{}}
\textbf{STG-KD \cite{pan2020spatio}:} A woman is cooking. \\ 
\textbf{Ours:} A person is cutting a piece of meat. \\
\textbf{GT:} A person is cutting mushroom.
\endtabular
}
\end{center}
\end{subfigure}
\begin{subfigure}{0.33\textwidth}
\begin{center}
% \fbox{\rule{0pt}{2in} \rule{.9\linewidth}{0pt}}
\includegraphics[height=0.4\linewidth,width=0.9\linewidth]{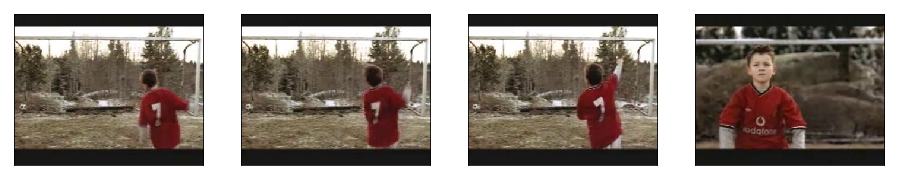}
\includegraphics[height=0.4\linewidth,width=0.9\linewidth]{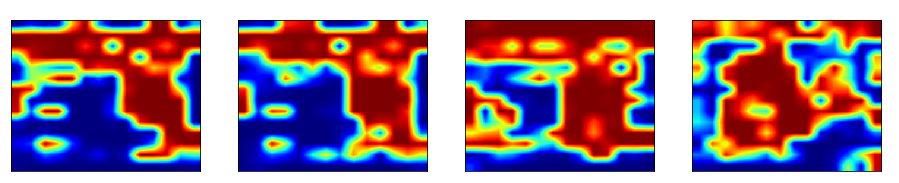}
\vspace{-0.2cm}
\caption{
\centering\label{fig:qual1}
\tabular[t]{@{}l@{}}
\textbf{STG-KD \cite{pan2020spatio}:} A boy kicks a goal. \\ 
\textbf{Ours:} A boy is kicking a soccer ball. \\
\textbf{GT:} A boy kicks a soccer ball.
\endtabular
} 
\end{center}
\end{subfigure}
\begin{subfigure}{0.33\textwidth}
\begin{center}
% \fbox{\rule{0pt}{2in} \rule{.9\linewidth}{0pt}}
\includegraphics[height=0.4\linewidth,width=0.9\linewidth]{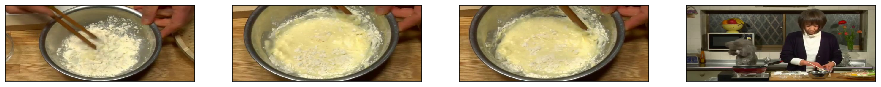}
\includegraphics[height=0.4\linewidth,width=0.9\linewidth]{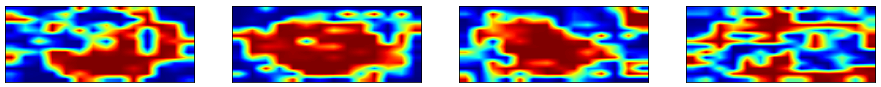}
\vspace{-0.2cm}
\caption{
\centering\label{fig:qual2}
\tabular[t]{@{}l@{}}
\textbf{STG-KD \cite{pan2020spatio}:} A person is cooking. \\ 
\textbf{Ours:} A woman is mixing ingredients. \\
\textbf{GT:} A woman is mixing water and flour.
\endtabular
} 
\end{center}
\end{subfigure}
\vspace{-0.2cm}
\caption{Qualitative visualization of CoSB branch's COSAM spatial attention on MSVD  video captioning dataset (best viewed in color)}
\vspace{-0.2cm}
\end{figure*}

\begin{figure*}[!ht]
\begin{subfigure}{0.5\textwidth}
\begin{center}
% \fbox{\rule{0pt}{2in} \rule{.9\linewidth}{0pt}}
\includegraphics[height=0.25\linewidth,width=0.9\linewidth]{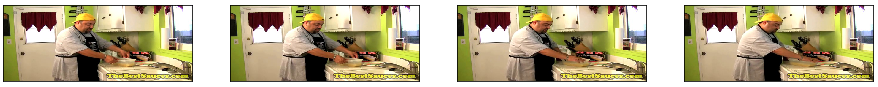}
\includegraphics[height=0.25\linewidth,width=0.9\linewidth]{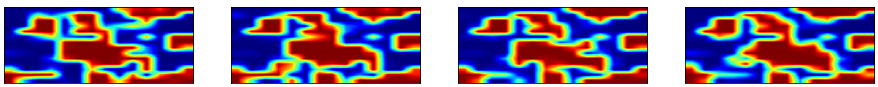}
\vspace{-0.2cm}
\caption{
\centering\label{fig:qual3}
\begin{tabular}{l}
\textbf{STG-KD \cite{pan2020spatio}:} A man is pouring pasta on to a container. \\ 
\textbf{Ours:} A man is putting a lid on a plastic container. \\
\textbf{GT:} A man puts a lid on a plastic container.
\end{tabular}
} 
\end{center}
\end{subfigure}
% \begin{figure}[!ht]
\begin{subfigure}{0.5\textwidth}
\begin{center}
\includegraphics[height=0.25\linewidth,width=0.9\linewidth]{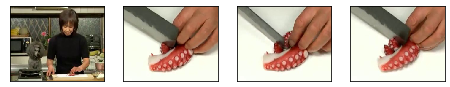}
\includegraphics[height=0.25\linewidth,width=0.9\linewidth]{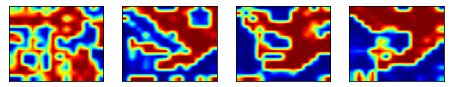}
\vspace{-0.2cm}
\caption{
\centering \label{fig:qual4}
\begin{tabular}{l}
\textbf{STG-KD \cite{pan2020spatio}:} A woman is slicing carrot.\\
\textbf{Ours:} A woman is slicing octopus. \\
\textbf{GT:} A woman is slicing octopus.
\end{tabular}
} 
\end{center}
\end{subfigure}
\vspace{-0.2cm}
\caption{Qualitative visualization of CoSB branch's COSAM spatial attention on MSVD video captioning  dataset (best viewed in color)}
\vspace{-0.4cm}
\end{figure*}

\subsubsection{Comparison with state-of-the-art methods}

We compare our model with recent state-of-the-art methods: {SA-LSTM} (\cite{wang2018reconstruction}), {M3} (\cite{wang2018m3}), {RecNet} (\cite{wang2018reconstruction}), {PickNet} (\cite{chen2018less}), {MARN} (\cite{pei2019memory}), {POS+VCT} (\cite{hou2019joint}), {ORG-TRL} (\cite{zhang2020object}), {OA-BGT} (\cite{zhang2019object}), {STG-KD} (\cite{pan2020spatio}).

Table \ref{tab:sota} shows the comparison of our model with the state-of-the-art methods.
Table \ref{tab:sota} is split into three logical groups that signifies different line of works in the video-captioning literature. First group (POS+VCT (\cite{hou2019joint}), ORG-TRL (\cite{zhang2020object})) focuses on improving decoder in their video captioning model by directly optimizing the decoder with \textit{$<$POS$>$} tags and pre-trained BERT (\cite{devlin2018bert}) modules. The second group ({OA-BTG} (\cite{zhang2019object}), STG-KD (\cite{pan2020spatio})) employ a pre-trained object detector to introduce the concept of objects inside the model. The third group (RecNet ( \cite{wang2018reconstruction}), PickNet (\cite{chen2018less}), MARN (\cite{pei2019memory}), SA-LSTM (\cite{wang2018reconstruction}), M3 (\cite{wang2018m3})) focuses on encoders without the help of external pre-trained models like object detectors / optical flow estimators. Our model belongs to third category where we focus on getting a better encoder representation without using external pre-trained models and dependencies. Note that we follow the standard procedure in the literature (\cite{pei2019memory}) to not compare with models based on reinforcement Learning (RL) (\cite{wang2019controllable}) techniques.

From the Table \ref{tab:sota}, in {MSRVTT} dataset, we can see that our model gets competitive performance in {B@4, ROUGE\_L} measures.  Specifically, our model performs 1 and 0.3 points better than its most comparable method {MARN} in {B@4} and { ROUGE\_L} respectively. Further, our model performs 1.2 and 2.4 points over PickNet (\cite{chen2018less}) in {ROUGE\_L} and {CIDEr} metrics respectively. We hypothesize that the moderate performance of our model in MSR-VTT is due to the high number of discontinuous clips in these videos, thus it may be difficult for COSAM to determine salient regions.

In {MSVD} dataset, our model outperforms on 3 metrics in the third logical group: METEOR, ROUGE\_L and CIDEr showing the effectiveness of combination of GSB and CoSB branches. We achieve state-of-the-art CIDEr score outperforming even the models that use external object detectors (\cite{chen2019mmdetection}) and those that focuses on both encoder and decoder (\cite{zhang2020object}). CIDEr being a metric that is better correlated with human judgement, this performance improvement shows that our model is able to generalize to test set and generate better captions.

\begin{table}[!ht]
\vspace{-0.3cm}
\begin{center}
\begin{tabular}{|c|c|c|c|c|c|c|}
\hline
\multicolumn{3}{|c|}{Components} & \multirow{3}{*}{B@4} & 
\multirow{3}{*}{M} & \multirow{3}{*}{R} & \multirow{3}{*}{C} \\
\cline{1-3}
% \hline
\multirow{2}{*}{GSB}&\multicolumn{2}{|c|}{CoSB}&&&& \\
\cline{2-3}
&COSAM&SRIM&&&& \\
\hline
\cmark & \xmark & \xmark &48.0&34.5&71.2&93.3\\
\hline
\xmark & \cmark& \cmark & 38.3&28.3&64.7&49.1\\
\hline
\cmark& \cmark& \xmark& 50.4&34.4&71.7&92.7 \\
\hline
\cmark& \xmark& \cmark& 49.2& 35.2&71.9& 92.7 \\
\hline
\cmark& \cmark& \cmark& 49.4&35.0&71.5&96.9\\
\hline
\cmark& \cmark& \cmark(m) & \textbf{50.7}&\textbf{35.3}&\textbf{72.1}&\textbf{97.8} \\
\hline
\end{tabular}
\end{center}
\vspace{-0.5cm}
\caption{Ablation study of the proposed model on MSVD dataset. Here, B@4 = BLUE@4, M = METEOR, R = ROUGE\_L, C = CIDEr. 'm' in SRIM column indicates masked multihead self-attention.\label{tab:ablation}}
\vspace{-0.5cm}
\end{table}

\vspace{-0.2cm}
\subsubsection{Qualitative Analysis}
In this section, we illustrate some of the interesting qualitative visualization from our model showing its ability to localize the important task-specific salient regions to aid video captioning.
As our model is similar to \cite{pan2020spatio}\footnote{We reproduce and train the model on our own for producing these results.} but without using external object detector,  we also provide captions from \cite{pan2020spatio} in the figures to have a relative comparison. In the qualitative visualizations (Fig. \ref{fig:qual0}, \ref{fig:qual1}, \ref{fig:qual2}, \ref{fig:qual3} and \ref{fig:qual4}), we show the visualizations of the spatial mask from COSAM module of CoSB branch. From the visualizations, it is evident from the masks that our model is able to attend key salient regions such as objects, persons in the video. Specifically, in Fig. \ref{fig:qual0}, our model attempts to generate more detailed captions by adding a quantifier like "a piece of". Similarly, in Fig. \ref{fig:qual2}, "mixing ingredients" is more detailed about the particular action than "cooking". In Fig. \ref{fig:qual3}, our model is able to capture the "closing`` action between the  person and the box lid, as well as it is able to generate a caption giving specific details about "plastic container``. Further, in Fig. \ref{fig:qual1}, our model is able to attend to the cross bar of the goal post which is an important cue to showcase that the boy is hitting a "soccer ball". We notice a key observation that our model is better at classifying/recognizing objects that are important for captioning tasks. For example, Fig. \ref{fig:qual4} shows that our model is correctly able to classify "octopus", where as \cite{pan2020spatio} classifies it as "carrot". We hypothesize that object detectors may face difficulty in detecting and recognizing such uncommon classes that are not present in their training dataset. Hence, letting the model to figure out the salient regions without biasing it with an external object detector is a potent approach.

\vspace{-0.1cm}
\subsubsection{Ablation studies \label{sec:ablation_caption}}

In this section, we perform ablation studies of different components of the CoSB branch. The ablation study results are shown in Table \ref{tab:ablation}.
In the first experiment (Row 1), we use a simple model by having only the GSB with the cross entropy loss and show the performance. 
In the next experiment (Row 2), we train only {CoSB} with the cross entropy loss which performs inferior to GSB-only model. It may be due to the use of only 2D intermediate features without temporal context. Further, from the third experiment, along with GSB, we start integrating the components of CoSB one by one. Row 3 shows the performance of our model with GSB and COSAM. This variant improves the GSB-only performance by 2.4, 0.5 points in B@4 and ROUGE\_L metrics respectively. Similarly, we train GSB and SRIM (Row 4) to showcase relevance of spatial interactions which results in improved in METEOR and ROUGE\_L.  
Row 5 shows the performance of the model with GSB+COSAM+SRIM. It can be seen that this combination improves CIDEr metric significantly by 3.6 points over GSB-only model. In the next experiment (Row 6), we modify SRIM to restrict interaction of salient-region features (i.e., use a mask in multi-head self-attention) to be only within adjacent frames. We observe that this masked multi-head self-attention improves the model's performance. Specifically, it improves 1.3, 0.9 points on B@4 and CIDEr metrics over the model where SRIM has all-to-all self attention. We hypothesize that constraining salient region interaction only to adjacent frames may avoid noisy feature interaction between frames with higher time gap, thus improving the performance.

\begin{table*}[!ht]
    \vspace{-0.3cm}
    \centering
    \begin{tabular}{|l|c|c|c|c|c|c|c|c|}
       \hline
      \multirow{2}{*}{Backbone} &  \multirow{2}{*}{COSAM?}  & \multirow{2}{*}{\makecell{temporal\\ modeling?}} & \multirow{2}{*}{\makecell{\#params\\ (M)}} & \multirow{2}{*}{\makecell{\#Flops (G)}} &  \multicolumn{2}{|c|}{HMDB51} & \multicolumn{2}{|c|}{UCF101} \\\cline{6-9}
       & & & & & Top-1\% & Top-3\% & Top-1\% & Top-3\%  \\\hline
       \makecell[l]{ResNeXt101\\(\cite{kalfaoglu2020late})} & \xmark & LSTM & 47.6 & 38.64 & 73.68 & 87.46 & 93.90 & 98.05 \\\hline
       ResNeXt101 & \cmark & LSTM & 48.41 & 38.77 & \textbf{75.16} & \textbf{89.22} & \textbf{94.59} & \textbf{98.52}  \\\hline\hline
       \makecell[l]{ResNeXt101\\(\cite{kalfaoglu2020late})} & \xmark & BERT & 47.4 & 38.37 & 76.08 & 90.46 & 95.50 & 98.23   \\\hline
       ResNeXt101 & \cmark & BERT & 48.21 & 38.49 & \textbf{77.52} & \textbf{92.55} &  \textbf{95.96} & \textbf{98.84}  \\\hline\hline
       \makecell[l]{I3D\\(\cite{kalfaoglu2020late})} & \xmark & BERT & 13.57 & 110.6 & 68.63 & 87.78 & 92.50 & 98.26 \\\hline
       I3D & \cmark & BERT & 14.23 & 110.7 & {69.38} & {87.95} & 93.05 & 98.63\\\hline
    %   \hline
    %   \makecell[l]{R(2+1)D\\(\cite{kalfaoglu2020late})} & \xmark & BERT & & & & \\\hline
    %   R(2+1)D & \cmark & BERT & & & & \\\hline
    \end{tabular}
    \vspace{-0.2cm}
    \caption{The performance comparison of single stream RGB model from \cite{kalfaoglu2020late} with and without COSAM layer.}  
    \label{tab:compare_videoclassify}
    \vspace{-0.3cm}
\end{table*}

\subsection{Video action classification}
\label{sec:experiments_videoclassify}

In this section, we present the experimental results of applying COSAM to video action classification task along with comparison to the baseline architecture. 

\subsubsection{Datasets}

We conduct the experiments on two action classification datasets (HMDB51 (\cite{kuehne2011hmdb}) and UCF101 (\cite{soomro2012ucf101})) in the literature.

\paragraph{UCF101} This dataset includes 13220 videos amounting to more than 27 hours of video data, belonging to 101 action classes. 
This dataset consists of realistic user uploaded videos with minimum length of 28 frames and average length of 180 frames per video containing scenes with cluttered background and camera ego-motion. The 101 action classes of the dataset can be broadly clustered into five types namely: 1)  Human-Object Interaction, 2) Body-Motion Only, 3) Human-Human Interaction, 4) Playing Musical Instruments, and 5) Sports.

\paragraph{HMDB51} The dataset consists of 51 action categories, each containing $\sim$101 clips for a total of 6,766 video clips extracted from
movie scenes and YouTube. 
The videos contain various nuisance factors such as camera motion, viewpoint, pose variations, poor video quality and partial occlusions that may hinder action classification. 
Every video in this dataset has a minimum of 18 frames. The 51 action categories of the dataset can be divided into five types namely: 1) General facial actions,
2) Facial actions with object manipulation, 3) General body movements, 4) Body movements with object interaction, and 5) Body movements for human interactions.

\vspace{-0.1cm}
\subsubsection{Evaluation metrics}

The datasets are organized into three splits containing  predefined number of training and testing videos. Every UCF101 split consists of $9.5K$
training videos, where as an HMDB51 split contains $3.7K$ training videos.
To calculate the final performance of the model, we follow the standard evaluation procedure followed in the baseline (\cite{kalfaoglu2020late}) by finding the mean top-1, top-3 accuracy across three splits and report the results. 

\begin{table*}[!ht]
% \vspace{-0.1cm}
    \centering
    \begin{tabular}{|c|l|c|c|c|} \hline
    \multirow{9}{*}{\rotatebox[origin=l]{90}{Two-stream\quad\quad}} & \textbf{Method} & \textbf{use flow?} & \textbf{HMDB51} & \textbf{UCF101} \\\hline
      & TwoStream (\cite{simonyan2014two}) & \cmark & 59.40 & 88.00 \\\cline{2-5}
     & TwoStream Fusion + IDT (\cite{feichtenhofer2016convolutional}) & \cmark & 69.20 & 93.50 \\\cline{2-5}
     & ActionVlad + IDT (\cite{girdhar2017actionvlad}) & \cmark & 69.80 & 93.60 \\\cline{2-5}
     & Temp. Segment (\cite{wang2018temporal}) & \cmark & 71.00 & 94.90 \\\cline{2-5}
     & R(2+1)D (\cite{tran2018closer}) & \cmark & 78.70 & 97.30 \\\cline{2-5}  
    &  I3D (\cite{carreira2017quo}) & \cmark & 80.90 & 97.80 \\\cline{2-5}
    & MARS + RGB + Flow (\cite{crasto2019mars}) & \cmark & 80.90 & 98.10 \\\cline{2-5}
    & BubbleNet (\cite{lo2020bubblenet}) &  \cmark & 82.6 & 97.2\\\cline{2-5}
    & ResNeXt101 BERT (\cite{kalfaoglu2020late}) & \cmark & 83.55 & 97.87 \\\hline\hline
    \multirow{6}{*}{\rotatebox[origin=l]{90}{Single-stream\quad}} & IDT (\cite{wang2013action}) & \xmark & 61.70 & - \\\cline{2-5}
    & R(2+1)D (\cite{tran2018closer}) & \xmark & 74.50 & 96.80 \\\cline{2-5}
    &  MARS + RGB (\cite{crasto2019mars})  & \xmark & 73.10 & 95.60 \\\cline{2-5}
    & TemporalShift (\cite{lin2019temporal}) & \xmark & 73.50 & 95.90 \\\cline{2-5}
    & ResNeXt101 BERT (\cite{kalfaoglu2020late}) & \xmark & 76.08 & 94.59 \\\cline{2-5}
    & \textbf{ResNeXt101 + COSAM + BERT (ours)} & \xmark & 77.52 & 95.96 \\\hline
    \end{tabular}
    \vspace{-0.2cm}
    \caption{State-of-the-art performance comparison of deep models for video action classification task.}  
    \label{tab:sota_videoclassify}
    \vspace{-0.4cm}
\end{table*}

\vspace{-0.1cm}
\subsubsection{Implementation details}
We implement the network architecture using PyTorch framework (\cite{paszke2019pytorch}). We keep the training settings same as the baseline method (\cite{kalfaoglu2020late})\footnote{https://github.com/artest08/LateTemporalModeling3DCNN}. We utilize AdamW optimizer with $\beta1 = 0.9$, $\beta2 = 0.999$ and weight decay = $1e^{-3}$. During training, the following hyper-parameters are used: the video frames are resized to have  height = $112$, width = $112$, 64 frames are uniformly sampled from each video, batch size = 8 for ResNeXt101 and 4 for I3D, initial learning rate $10^{-5}$, we drop the learning rate by multiplying it with 0.1 when the loss plateaus after an epoch, 
The linear projection layers of BERT, the classification token (\textit{CLS}) and the learned positional embeddings are initialized with the zero-mean normal distribution with standard deviation of 0.02. 

\vspace{-0.1cm}
\subsubsection{Comparison with baseline method}

In this section, we illustrate the performance comparison between the baseline 3D CNN model (\cite{kalfaoglu2020late}) and COSAM-incorporated model. 
The experimental results are shown in the Table \ref{tab:compare_videoclassify}. Incorporating COSAM module into the ResNeXt-101 feature extractor (Tab. \ref{tab:compare_videoclassify} row 2) with LSTM as temporal modeling scheme improves the baseline by +1.5\% and +0.7\% top-1 accuracy on HMDB51 and UCF101 datasets respectively. Further, the COSAM-based model with ResNeXt-101 feature extractor and BERT pooling layer (Tab. \ref{tab:compare_videoclassify} row 4) performs better than its baseline model by +1.4\% \& +0.5\% on HMDB51 and UCF101 datasets respectively. Notably, the models with BERT layer outperform the models with LSTM temporal modeling layer as shown in \cite{kalfaoglu2020late}. From the performance improvements of COSAM models as shown in Tab. \ref{tab:compare_videoclassify},  it is evident that COSAM is able to generalize irrespective of the temporal modeling scheme (LSTM or BERT). 
Next, we incorporate COSAM into I3D feature extractor along with BERT temporal modeling layer to evaluate the generalization of COSAM across different feature extractors. As shown in the Table \ref{tab:compare_videoclassify}, I3D + COSAM model performs +0.75\% and +0.5\% better on HMDB51 and UCF101 top-1 accuracy respectively. The improvements on two different network architectures shows that COSAM is able to generalize the performance improvements and have potential to be applied across various model architectures.

\begin{figure*}[!ht]
    \centering
    \includegraphics[width=\textwidth]{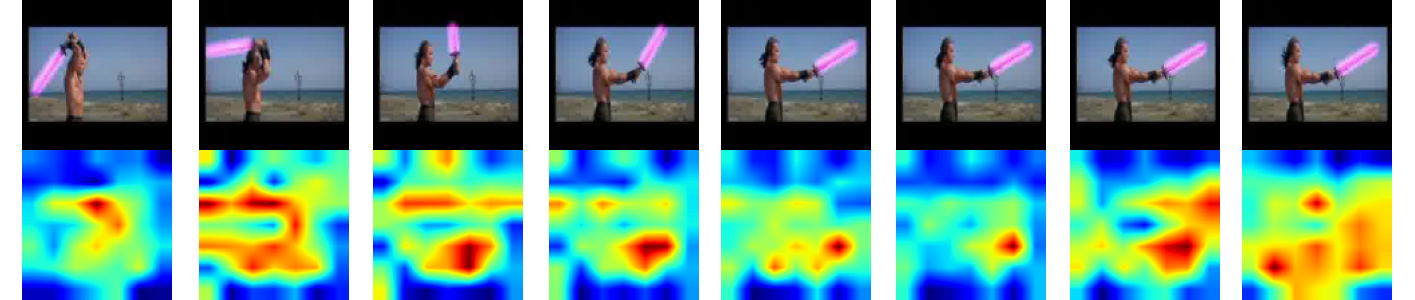}
    \vspace{-0.6cm}
    \caption{Visualization of COSAM spatial attention output for ``Sword Exercise'' class from HMDB51 dataset}
    \label{fig:videoclassify_qual0}
    \vspace{-0.2cm}
\end{figure*}

\begin{figure*}[!ht]
    \centering
    \includegraphics[width=\textwidth]{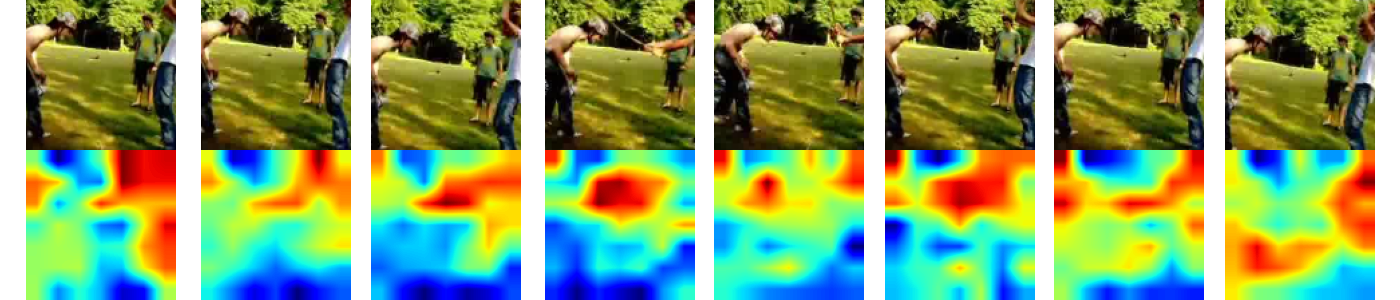}
    \vspace{-0.6cm}
    \caption{Visualization of COSAM spatial attention output for ``Hit'' class from HMDB51 dataset}
    \label{fig:videoclassify_qual1}
    \vspace{-0.2cm}
\end{figure*}

\begin{figure*}[!ht]
    \centering
    \includegraphics[width=\textwidth]{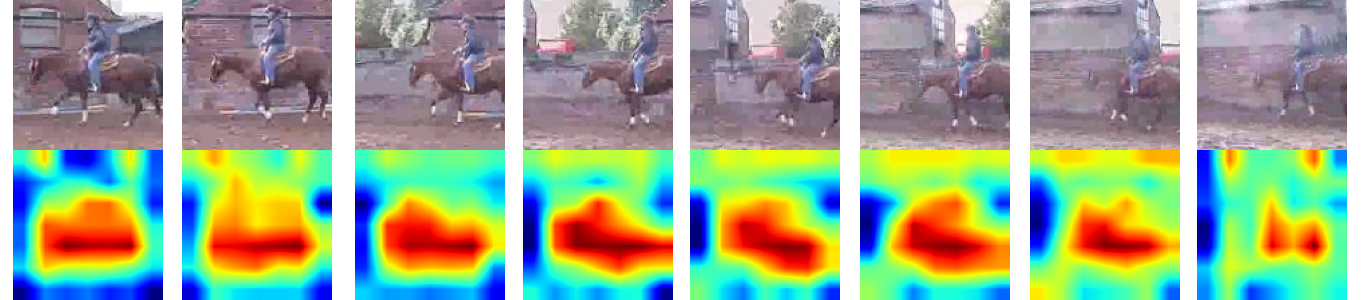}
    \vspace{-0.6cm}
    \caption{Visualization of COSAM spatial attention output for ``Horse Riding'' class from UCF101 dataset}
    \label{fig:videoclassify_qual2}
    \vspace{-0.2cm}
\end{figure*}

\begin{figure*}[!ht]
    \centering
    \includegraphics[width=\textwidth]{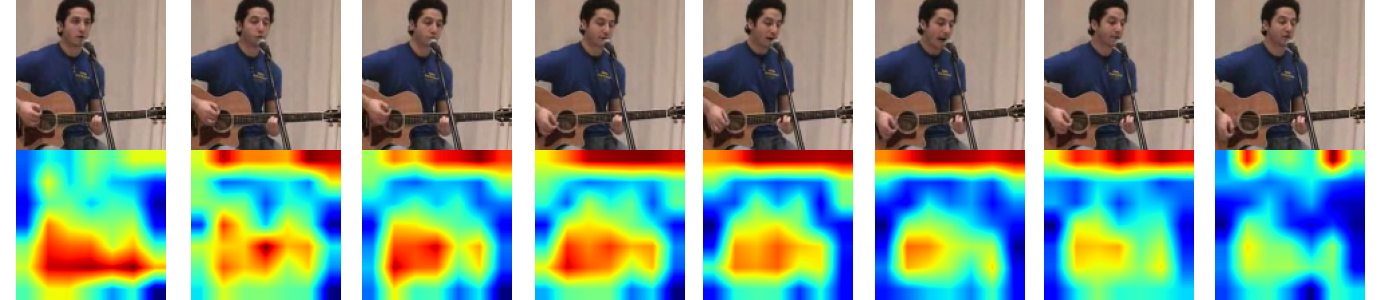}
    \vspace{-0.6cm}
    \caption{Visualization of COSAM spatial attention output for ``Playing Guitar'' class from UCF101 dataset}
    \label{fig:videoclassify_qual3}
    \vspace{-0.6cm}
\end{figure*}

\subsubsection{Comparison with the state-of-the-art methods}

We compare the COSAM-based model with the following state-of-the-art methods with single-stream models (IDT (\cite{wang2013action}), R(2+1)D (\cite{tran2018closer}), TemporalShift (\cite{lin2019temporal}), ResNeXt101 BERT (\cite{kalfaoglu2020late})) and models with two-streams (TwoStream (\cite{simonyan2014two}), TwoStream Fusion + IDT (\cite{feichtenhofer2016convolutional}), ActionVlad + IDT (\cite{girdhar2017actionvlad}), Temp. Segment (\cite{wang2018temporal}), R(2+1)D (\cite{tran2018closer}), I3D (\cite{carreira2017quo}), , MARS + RGB + Flow (\cite{crasto2019mars}), BubbleNet (\cite{lo2020bubblenet}), ResNeXt101 BERT (\cite{kalfaoglu2020late})).

The comparison with the state-of-the-art methods are shown in Table \ref{tab:sota_videoclassify}. Table \ref{tab:sota_videoclassify} is split into two logical categories namely: 1) \textit{Single-stream models} that use only the RGB frames of the videos as input, 2) \textit{Two-stream models} that use both RGB and optical flow inputs. Overall, the two-stream methods perform superior to the single-stream methods due to the explicit motion cues provided in terms of optical flow. However, they suffer from the disadvantage of high latency due to the need of optical flow estimation between video frames. In our work, we are motivated to improve the performance single-stream approach by incorporating COSAM into the model and thus we compare our method with single-stream approaches. In the single-stream category with only appearance input (RGB frames), our COSAM-based model (ResNeXt101+COSAM+BERT) performs superior to the state-of-the-art methods with only RGB input. Specifically, our method performs better than \cite{kalfaoglu2020late} with +1.44\% and +1.37\% Top-1 accuracy on HMDB51 and UCF101 datasets respectively. Further, our method shows significant performance improvement over TemporalShift (\cite{lin2019temporal}) by +4.02\%, MARS+RGB (\cite{crasto2019mars}) by +4.42\%, R(2+1)D (\cite{tran2018closer}) by +3.02\% respectively on HMDB51 dataset. COSAM can be considered as orthogonal approach to these methods and may be combined with some of the methods to improve the performance further.

\subsubsection{Qualitative Analysis}

In this section, we illustrate some of the interesting visualizations (Fig. \ref{fig:videoclassify_qual0}, \ref{fig:videoclassify_qual1}, \ref{fig:videoclassify_qual2} \& \ref{fig:videoclassify_qual3}) of COSAM spatial attention maps from the COSAM-based model with ResNeXt-101 backbone and BERT temporal modeling layer. 

In general, we observe that COSAM is able to attend to the salient object regions needed for the task (\textit{i.e.}, video action classification) and refine the features by suppressing the background noisy features. For instance, in Fig. \ref{fig:videoclassify_qual0}, to classify the class "Sword Exercise", COSAM is able to focus on the regions that the object (\textit{i.e.}, sword) is present in the image. Further, it is able to attend on the same salient regions coherently across all the video frames. Next, in Fig. \ref{fig:videoclassify_qual1}, the task-specific salient objects such as persons and baseball bat are captured in the attention map for the class ``Hit''. COSAM aids in automatically attending  to the objects that lie in tail distribution such as baseball bat which may not be detected by pre-trained object detectors. Next, Fig. \ref{fig:videoclassify_qual2} shows that COSAM is able to capture the person and horse regions to classify the video frames into ``Horse riding'' category. Note that the salient regions are detected regardless of partial occlusion and color ambiguity. Finally, Fig. \ref{fig:videoclassify_qual3} illustrates the ability of COSAM to focus on key semantic object (Guitar) needed to classify the action (Playing guitar) with high confidence than the other objects (i.e., Person) present in the video.

\vspace{-0.2cm}
\section{Conclusion and Future work}
\label{sec:conclusion}

In this work, we presented a generic co-segmentation based attention module (COSAM) that can be plugged-in within any backbone network. COSAM enables the model to automatically learn to attend to task-specific  salient object regions for the underlying task in an end-to-end manner. 
To illustrate the applicability of COSAM in various video-based computer vision tasks, we have incorporated it into three major tasks namely: 1) Video-based person re-ID, 2) Video captioning \& 3) Video action classification. Through a comprehensive set of empirical experiments, we have shown that inclusion of COSAM consistently improves the performance of the underlying baseline models with an added advantage of interpretable visual results. Further, we have shown a set of qualitative visualizations demonstrating the capability of COSAM to attend to salient regions for the underlying task. Given the generic nature of the COSAM module, its effectiveness on other video-based tasks such as visual tracking, video object segmentation can be pursued as a possible future work. 

\vspace{-0.3cm}
\section*{Acknowledgments}
\vspace{-0.1cm}
This work is supported by grants from PM’s fellowship for Doctoral Research (SERB, India) \& Google PhD Fellowship to Arulkumar Subramaniam.

\vspace{-0.3cm}
\bibliographystyle{model2-names}
\bibliography{refs}

% \section*{Supplementary Material}

\end{document}